\pdfoutput=1
\documentclass[11pt]{article}

\usepackage[]{emnlp2021}

\usepackage{times}
\usepackage{latexsym}

\usepackage[T1]{fontenc}
\usepackage[utf8]{inputenc}

\usepackage{amsfonts}
\usepackage{microtype}
\usepackage{booktabs}
\usepackage{multirow}
\usepackage{tabularx}
\usepackage{graphics}
\usepackage{adjustbox} 
\usepackage{subfigure}
\usepackage{amsmath}
\usepackage{hyperref}
\usepackage{caption}
\usepackage{graphicx}
\usepackage{algorithm}
\usepackage[noend]{algpseudocode}

\newcommand{\xhdr}[1]{{\noindent\bfseries #1}.}
\def\Snospace~{\S{}} % If i don't add this overleaf complains

\newcommand{\OR}{$\mathcal{M}_{\texttt{N}}$}
\newcommand{\RI}{$\mathcal{M}_{\texttt{1}}$}
\newcommand{\RII}{$\mathcal{M}_{\texttt{2}}$}
\newcommand{\RIII}{$\mathcal{M}_{\texttt{3}}$}
\newcommand{\RIV}{$\mathcal{M}_{\texttt{4}}$}
\newcommand{\RC}{$\mathcal{M}_{\texttt{UG}}$}
\newcommand{\RU}{$\mathcal{M}_{\texttt{UF}}$}
\newcommand{\RT}{$\mathcal{M}_{\texttt{RI}}$}
\newcommand{\NP}{$\mathcal{M}_{\texttt{NP}}$}
\newcommand{\RV}{$\mathcal{M}_{\texttt{512}}$}

\title{Masked Language Modeling and the Distributional Hypothesis:\\ Order Word Matters Pre-training for Little} %Transformers Order Word Matters Very Little For}
\author{Koustuv Sinha$^\dagger{}^\ddagger$ \quad Robin Jia$^\dagger$ \quad Dieuwke Hupkes$^\dagger$ \quad Joelle Pineau$^\dagger{}^\ddagger$ \AND Adina Williams$^\dagger$ \quad Douwe Kiela$^\dagger$\\
$^\dagger$ Facebook AI Research; $^\ddagger$ McGill University / Mila - Quebec AI\\
  \texttt{\{koustuvs,adinawilliams,dkiela\}@fb.com} \\}
\begin{document}
\maketitle
\begin{abstract}
A possible explanation for the impressive performance of masked language model (MLM) pre-training is that such models have learned to represent the syntactic structures prevalent in classical NLP pipelines.
%have learned a great deal about linguistic structure. %, and there is indeed some evidence to support this in prior work.
In this paper, we propose a different explanation:
MLMs succeed on downstream tasks mostly due to their ability to model higher-order word co-occurrence statistics.
To demonstrate this, we pre-train MLMs on sentences with randomly shuffled word order, and we show that these models still achieve high accuracy after fine-tuning on many downstream tasks - including tasks specifically designed to be challenging for models that ignore word order.
Our models also perform surprisingly well according to some parametric syntactic probes, indicating possible deficiencies in how we test representations for syntactic information. 
Overall, our results show that purely distributional information largely explains the success of pre-training,
and they underscore the importance of curating challenging evaluation datasets that require deeper linguistic knowledge.
\end{abstract}

\section{Introduction}

The field of natural language processing (NLP) has become dominated by the pretrain-and-finetune paradigm, where we first obtain a good parametric \emph{prior} in order to subsequently model downstream tasks accurately. 
In particular, masked language model (MLM) pre-training, as epitomized by BERT \citep{devlin-etal-2019-bert}, has proven wildly successful, although the precise reason for this success has remained unclear.
On one hand, we can view BERT as the newest in a long line of NLP techniques  \cite{deerwester1990indexing, landauer1997solution, collobert2008unified, mikolov2013, peters-etal-2018-deep} that exploit the well-known distributional hypothesis~\cite{harris1954distributional}.\footnote{One might even argue that BERT is not actually all that different from earlier distributional models like word2vec~\cite{mikolov2013}, see \autoref{app:bert-is-word2vec}.} 
On the other hand, it has been claimed that BERT ``rediscovers the classical NLP pipeline''~\cite{tenney-etal-2019-bert}, suggesting that it has learned ``the types of syntactic and semantic abstractions traditionally believed necessary for language processing'' %``the kind of abstractions that we intuitively believe are important for representing natural language'' 
rather than ``simply modeling complex co-occurrence statistics''~(ibid. p.1). % KS: i'm not sure what was the citation here % \citep[p. 4593]{}.

In this work, we aim to uncover how much of MLM's success comes from learning simple distributional information, as opposed to grammatical abstractions \citep{tenney-etal-2019-bert,manning2020emergent}. %``the types of syntactic and semantic abstractions traditionally believed necessary for language processing'' \cite{tenney2019bert, manning2020emergent}. %[citeTenney]. 
We disentangle these two hypotheses by measuring the effect of removing word order information during pre-training:
%DK: removed this because people will get confused by "local"
%while %distributional statistics
%local distributional co-occurrences
%are not sensitive to word order changes,
any sophisticated (English) NLP pipeline would presumably depend on the syntactic information conveyed by the order of words.
Surprisingly, we find that most of MLM's high performance can in fact be explained by the ``distributional prior'' rather than its ability to replicate the classical NLP pipeline.

Concretely, we pre-train MLMs~(RoBERTa, \citealt{liu2019b}) on various corpora with permuted word order while preserving some degree of distributional information, and examine their downstream performance. 
%In our main experiments, we pre-train models on 
%various permuted corpora , by randomly shuffling $n$-grams within the sentence (where $n \in \{1,2,3,4\}$). 
We also experiment with training MLMs without positional embeddings, making them entirely order agnostic,
and with training on a corpus sampled from the source corpus's %uniform or 
unigram distribution%, removing both distributional and word order information
. We then evaluate these ``permuted'' models in a wide range of settings and compare with regularly-pre-trained models. 

We demonstrate that pre-training on permuted data has surprisingly little effect on downstream task performance after fine-tuning (on non-shuffled training data). 
%This is partially explained by inductive biases in the model itself, as illustrated by the baselines performing reasonably well.  Those are vastly outperformed, however, by permuted models with distributional information preserved at the sentence level, which perform remarkably close to regular MLM pre-training.
It has recently been found that MLMs are quite robust to permuting downstream test data~\cite{sinha2020b, pham2020, gupta-etal-2021-bert} and even do quite well using permuted ``unnatural'' downstream train data \cite{sinha2020b,gupta-etal-2021-bert}. Here, we show that downstream performance for ``unnatural language pre-training'' is much closer to standard MLM pre-training than one might expect.

In an effort to shed light on these findings, we experiment with various probing tasks. We verify via non-parametric probes that the permutations do in fact make the model worse at syntax-dependent tasks. However, just like on the downstream fine-tuning tasks, permuted models perform well on parametric syntactic probes, in some cases almost matching the unpermuted model's performance, which is quite surprising given how important word order is crosslinguistically (\citealt{greenberg1963some, dryer1992greenbergian, cinque1999adverbs}, i.a.).
% [citeDryer?Greenberg].
%This again suggests that MLM's success in downstream tasks is mostly explained by it having learned higher-order distributional statistics that make for a useful prior.

% Depending on one's position in NLP's great debates, the reader might draw different conclusions from these findings. 
% One might argue that our downstream tasks are flawed evaluations, and that we need to examine models with examples that truly test strong generalization and compositionality. 
% Alternatively, one could argue that human language understanding simply depends for a large part on the structure (i.e., order) of the \emph{world}. 
% Or, perhaps we are just not focusing enough on the tail of the distribution, where true syntactic capabilities would certainly be borne out. 
% While these results may seem disappointing for NLP at large, this work also offers a message of hope and renewed purpose: can we design tasks that require more sophisticated reasoning, true compositionality, robustness against adversaries and strong human-like generalization? 
% Can we design better experimental protocols for testing the phenomena we really care about? 
% Can we be more careful with our baselines? 
% This work is meant to deepen our understanding of MLM pre-training, and through this, orient future work towards promising new research directions.

Our results can be interpreted in different ways. 
One could argue that our downstream and probing tasks are flawed, and that we need to examine models with examples that truly test strong generalization and compositionality.
Alternatively, one could argue that prior works have overstated the dependence of human language understanding on word order, and that human language understanding depends less on the structure of the sentence and more on the structure of the \emph{world}, which can be inferred to a large extent from distributional information.
This work is meant to deepen our understanding of MLM pre-training and, through this, move us closer to finding out what is actually required for adequately modelling natural language.

%%% Previous intro ending
% Our contributions:

% \begin{itemize}
%     \item We find permuting word order while pre-training to have surprisingly low effect on downstream tasks (GLUE, PAWS).
%     \item We observe majority of word order information models acquire is learned during fine-tuning, with some tasks being more sensitive than others.
%     \item We observe that non-parametric measures are much better at distinguishing pre-trained models trained on permuted word order than parametric probing tasks.
% \end{itemize}

\section{Related Work}

\xhdr{Sensitivity to word order in NLU} Information order has been a topic of research in computational linguistics since \newcite{barzilay-lee-2004-catching} introduced the task of ranking sentence orders
%text-ordering task: algorithm has to select a maximally coherent sentence order from a set of candidate permutations
as an evaluation for language generation quality, an approach which was subsequently also used to evaluate readability and dialogue coherence \citep{barzilay-lapata-2008-modeling, laban-etal-2021-transformer}. 

More recently, several research groups have investigated information order for words rather than sentences as an evaluation of model humanlikeness. %\citet{sinha2020b}, \citet{pham2020}, and \citet{gupta-etal-2021-bert}.
\citet{sinha2020b} investigate the task of natural language inference (NLI) and find high accuracy on permuted examples for different Transformer and pre-Transformer era models, across English and Chinese datasets \cite{hu-etal-2020-ocnli}.
\citet{gupta-etal-2021-bert} use targeted permutations on RoBERTa-based models and show word order insensitivity across natural language inference (MNLI), paraphrase detection (QQP) and sentiment analysis tasks (SST-2). 
%RoBERTa has been found to assign high confidence to targeted permutations, even with drastic changes in word order.
\citet{pham2020} show insensitivity on a larger set of tasks, including the entire GLUE benchmark, and find that certain tasks in GLUE, such as CoLA and RTE are more sensitive to permutations than others.
\newcite{ettinger-2020-whatbertisnot} recently observed that BERT accuracy decreases for some word order perturbed examples, but not for others. 
In all these prior works, models were given access to normal word order at (pre-)training time, but not at fine-tuning or test time. It was not clear whether the model acquires enough information about word order during the fine-tuning step, or whether it is ingrained in the pre-trained model. 
% This could mean that models' achieve high accuracy on test datasets because the information about word order learned from standard pre-training outweighs what is learned from fine-tuning on permuted data.
In this work, we take these investigations a step further: we show that the word order information needed for downstream tasks does not need to be provided to the model during pre-training. Since models can learn whatever word order information they do need largely from fine-tuning alone, this likely suggests that our downstream tasks don't actually require much complex word order information in the first place (cf.,\ \citealt{glavas-vulic-2021-supervised}). 
% and show that training on word order permuted sentences only degrades test set accuracy somewhat - some important word order information can be learned solely from normal-order fine-tuning (cf.\ \citealt{howard-ruder-2018-universal}).

\xhdr{Randomization ablations}
% KS: I'm unsure how to write this section. @Douwe can you have a look?
% \cite{shen2020a}, chinna sankar, ettinger what bert is not 2020
% New section here
% Probably also want to cite Random Sentence Encoders/Reservoir Transformers and all the other "look randomness doesn't hurt as much as you'd think papers"?
% Do latent tree models identify meaningful:  
% \cite{shen2020a}
% SPINN paper Adina Tree Randomization paper
%
%- Deep image prior: https://arxiv.org/abs/1711.10925
%- Training BatchNorm and Only BatchNorm: On the Expressive Power of Random Features in CNNs: https://arxiv.org/abs/2003.00152
%- Exploring Randomly Wired Neural Networks for Image Recognition https://openaccess.thecvf.com/content_ICCV_2019/papers/Xie_Exploring_Randomly_Wired_Neural_Networks_for_Image_Recognition_ICCV_2019_paper.pdf
%
Random controls have been explored in a variety of prior work. 
\citet{wietingkiela2019} show that random sentence encoders are surprisingly powerful baselines.  \citet{gauthier-levy-2019-linking} use random sentence reordering to label some tasks as ``syntax-light'' making them more easily decodeable from images of the brain. 
\citet{shen2020a} show that entire layers of MLM transformers can be randomly initialized and kept frozen throughout training without detrimental effect and that those layers perform better on some probing tasks than their frozen counterparts.
Models have been found to be surprisingly robust to randomizing or cutting syntactic tree structures they were hoped to rely on~\cite{scheible2013cutting,williams2018latent}, and randomly permuting attention weights often induces only minimal changes in output \cite{jain-wallace-2019-attention}. In computer vision, it is well known that certain architectures constitute good ``deep image priors'' for fine-tuning~\cite{ulyanov2018deep} or pruning~\cite{frankle2020training}, and that even randomly wired networks can perform well at image recognition~\cite{xie2019exploring}. Here, we explore randomizing the data, rather than the model, to assess whether certain claims about which phenomena the model has learned are established in fact.

% Commenting this section out as per suggestion by Dieuwke

\xhdr{Synthetic pre-training} \citet{kataoka2021} found that pre-training on synthetically generated fractals for image classification is a very strong prior for subsequent fine-tuning on real image data. In language modeling, \citet{papadimitriou-jurafsky-2020-learning} train LSTMs \citep{hochreiter1997long} on non-linguistic data with latent structure such as MIDI music or Java code provides better test performance on downstream tasks than a randomly initialized model. They observe that even when there is no vocabulary overlap among source and target languages, LSTM language models leverage the latent hierarchical structure of the input to obtain better performance than a random, Zipfian corpus of the same vocabulary.

%by leveraging mathematical formula-driven fractals. They provide evidence that a Convolutional Neural Network (CNN) model, pre-trained with this large, automatically generated fractal data, can even surpass the accuracy of models trained on natural data, such as ImageNet.
%They observe that even when there is no vocabulary overlap among source and target languages, LSTM language models leverages the latent hierarchical structure of the input to obtain better performance than a random, Zipfian corpus of the same vocabulary. In our work, we go a step further to pre-train large Transformer language models on sentence randomized data, to find it has little effect on the downstream tasks.

% \xhdr{Randomness in Encoders}
% \cite{papadimitriou2020}
% Should also cite the CV paper https://arxiv.org/abs/2101.08515 (could actually lead the section with that, given how general that makes the paper look?)
% \cite{kataoka2021}

\xhdr{On the utility of probing tasks}
Many recent papers provide compelling evidence that BERT contains a surprising amount of syntax, semantics, and world knowledge \cite{giulianelli-etal-2018-hood,rogers-etal-2020-primer,lakretz-etal-2019-emergence,jumelet-etal-2019-analysing,jumelet-etal-2021-language}. Many of these works involve diagnostic classifiers \cite{hupkes2018visualisation} or \textit{parametric} probes, i.e.\ a function atop learned representations that is optimized to find linguistic information. How well the probe learns a given signal can be seen as a proxy for linguistic knowledge encoded in the representations. % \citep{hupkes2018visualisation}
However, the community is divided on many aspects of probing \citep{belinkov2021probing} including how complex probes should be.
Many prefer \textit{simple} linear probes over the complex ones \cite{alain2016understanding, hewitt2019structural, hall-maudslay-etal-2020-tale}. However, complex probes with strong representational capacity are able to extract the most information from representations \citep{voita-titov-2020-information, pimentel-etal-2020-information, hall-maudslay-etal-2020-tale}.
% On the one hand, proponents of \textit{simple} linear probes argue that having explicit, focused classifiers should provide better signal as the function itself is too weak on its own \cite{alain2016understanding, hewitt2019structural, hall-maudslay-etal-2020-tale}. On the other hand, proponents of \textit{complex} probes claim that having strong representational capacity of the probe is actually helpful for extracting the most information from a representation \cite{pimentel2020}. There are also strong calls from the community to use information theoretic approaches instead of accuracy \citep{voita-titov-2020-information, pimentel2020}, as well  harder \textit{tasks} for probing \cite{hall-maudslay-etal-2020-tale, pimentel-etal-2020-pareto}. 
Here, we follow \citet{pimentel-etal-2020-pareto} and use \textit{both} simple (linear) and complex (non-linear) models, as well as ``complex'' tasks (dependency parsing). 
As an alternative to parametric probes,  stimulus-based \textit{non-parametric} probing \cite{linzen-etal-2016-assessing, jumelet-hupkes-2018-language, marvin-linzen-2018-targeted, gulordava-etal-2018-colorless, warstadt-etal-2019-investigating, warstadt-etal-2020-blimp-benchmark, warstadt-etal-2020-learning, ettinger-2020-whatbertisnot,lakretz2021mechanisms} has been used to show that even without a learned probe, BERT can predict syntactic properties with high confidence~\cite{goldberga, wolf2019}.
%Stimulus-based probing is also sometimes called \textit{non-parametric}, since no additional parameters need to be learned to probe with this method. 
We use this class of non-parametric probes to investigate RoBERTa's ability to learn word order during pre-training.

% KS->Adina: do you want to take this section?
% \cite{pimentel2020, pimentel-etal-2020-pareto,hall-maudslay-etal-2020-tale}
% \cite{rogers2021}
% \cite{perez2021}

% \section{Background}

% Should serve as technical background of MLM and RoBERTa

% Can we re-use sentence superiority effect motivation?

\section{Approach}
\label{sec:experimental_setup}

%Before presenting our experiments on the impact of distributional prior in Masked Language Modeling (MLM) pre-training, 
We first describe the data generation and evaluation methodology used in this paper.
We use the RoBERTa (base) \cite{liu2019b} MLM architecture, due to its relative computational efficiency and good downstream task performance. We expect that other variants of MLMs would provide similar insights, given their similar characteristics.

\subsection{Models}

In all of our experiments, we use the original 16GB BookWiki corpus (the Toronto Books Corpus, \citealt{zhu2015aligning}, plus English Wikipedia) from \citet{liu2019b}.\footnote{We release the pre-trained RoBERTa models used in our experiments through the FairSeq repository:  \href{https://github.com/pytorch/fairseq/tree/master/examples/shuffled_word_order}{https://github.com/pytorch/fairseq/tree/master/examples /shuffled\_word\_order}.} We denote the model trained on the original, un-modified BookWiki corpus as \OR{} (for ``natural''). We use two types of word order randomization methods: permuting words at the sentence level, and resampling words at the corpus level.

\xhdr{Sentence word order permutation}
To investigate to what extent the performance of MLM pre-training is a consequence of distributional information, we construct a training corpus devoid of natural word order but preserving local distributional information. We construct word order-randomized versions of the BookWiki corpus, following the setup of \citet{sinha2020b}. %, where each sentence is stripped of its original word order. 
Concretely, given a sentence $S$ containing $N$ words, we permute the sentence using a seeded random function $\mathcal{F}_1$ such that no word can remain in its original position. In total, there exist $(N-1)!$ possible permutations of a given sentence. We randomly sample a single permutation per sentence, to keep the total dataset size similar to the original.
%the control case (the natural word order dataset).
% Hmm actually I double checked and we don't do that in the code.
% Similar to the construction of \citet{sinha2020b}, we keep the punctuations in their original position as these tokens are shown to be used by BERT as ``no-op" operations \cite{clark2019does} to focus substantial amount of attention, aiding in optimization.

% TODO: insert some math

We extend the permutation function $\mathcal{F}_1$ to a function $\mathcal{F}_n$ that preserves $n$-gram information. Specifically, given a sentence $S$ of length $N$ and $n$-gram value $n$, we sample a starting position $i$ for possible contiguous $n$-grams $\in \{0, N-n\}$ and convert the span $S[i,i+n]$ to a single token, to form $\hat{S}$, of length $\hat{N} = N - (n+1)$. We continue this process repeatedly (without using the previously created n-grams) until there exists no starting position for selecting a contiguous n-gram in $\hat{S}$. For example, given a sentence of length $N=6$, $\mathcal{F}_4$ will first convert one span of 4 tokens into a word, to have $\hat{S}$ consisting of three tokens (one conjoined token of 4 contiguous words, and two leftover words).
Then, the resulting sentence $\hat{S}$ is permuted using $\mathcal{F}_1$. We train RoBERTa models on four permutation variants of BookWiki corpus, \RI, \RII, \RIII, \RIV\ for each $n$-gram value $ \in {\{1,2,3,4\}}$. More details on the process, along with the pseudo code and sample quality, are provided in \autoref{app:data}.

% Concretely, we permute the word order of the sentence such that n-grams are kept unchanged in their relative word order.
% Since we sample a permutation of each class at random, certain percentage of relative word order remains intact in 2,3 and 4 gram permutation.
% In total, we construct four variants of BookWiki corpus of the same size: $R_1,R_2,R_3$ and $R_4$.

\xhdr{Corpus word order bootstrap resample}
The above permutations preserve higher order distributional information by keeping words from the same sentence together. However, we need a baseline to understand how a model would perform without such co-occurrence information. 
% As a  weak baseline, we consider a randomly initialized RoBERTa model. 
%A randomly initialized RoBERTa is a good baseline for seeing how far we can get using only the model architecture as the inductive bias. 
We construct a baseline, \RC{}, that captures word/subword information, without access to co-occurrence statistics. 
% In order to understand the effect of the distributional prior on its own, we need to isolate the inductive bias of the Transfomer model. 
To construct \RC, we sample unigrams from BookWiki according to their frequencies, while also treating named entities as unigrams. We leverage Spacy~\citep{spacy}\footnote{\href{https://spacy.io/}{https://spacy.io/}} to extract unigrams and named entities from the corpus, and construct \RC\ by drawing words from this set according to their frequency.
%Concretely, we first use Spacy's \citep{spacy}\footnote{\href{https://spacy.io/}{https://spacy.io/}} part-of-speech (POS)~tagger and named entity recognition (NER) parser to extract all POS tagged words and NER words (and their counts) from the source corpus (BookWiki), to form a word-to-frequency dictionary $\mathcal{W}$. For a phrase extracted by the NER parser, we remove the sub-tokens from the POS tag word set to maintain the same frequency of the tokens in BookWiki. Then, we replace the words in the original sentences $S_i \in$ BookWiki with words drawn from a weighted distribution of words (by frequency) from $\mathcal{W}$. 
This allows us to construct \RC\ such that it has exactly the same size as BookWiki but without any distributional (i.e. co-occurrence) information beyond the unigram frequency distribution. Our hypothesis is that any model pre-trained on this data will perform poorly, but it should provide a baseline for the limits on learning language of the inductive bias of the model in isolation.

\xhdr{Further baselines}
% We also train further model ablations with low distributional prior. Following the construction of corpus bootstrap resample, we train a model where words are drawn uniformly from BookWiki corpus, thus destroying the natural frequency distribution (\RU). 
To investigate what happens if a model has absolutely no notion of word order, we also experiment with pre-training RoBERTa on the original corpus without positional embeddings.
Concretely, we modify the RoBERTa architecture to remove the positional embeddings from the computation graph, and then proceed to pre-train on the natural order BookWiki corpus. 
We denote this model \NP. 
Finally, we consider a randomly initialized RoBERTa model \RT\, to observe the extent we can learn from each task with only the model's base inductive bias.
%TODO: motivation on why we need this baseline
%This baseline is provided with the correct distributional prior, however it is incapable of understanding word order. 
% is theoretically similar to $R_C$ where it also does not preserve distributional information of the tokens.

%F(pretraining)
%F(finetuning)

\xhdr{Pre-training details}
Each model $\in \{$\OR, \RI, \RII, \RIII, \RIV, \RC, \NP$\}$ is a RoBERTa-base model (12 layers, hidden size of 768, 12 attention heads, 125M parameters), trained for 100k updates using 8k batch-size, 20k warmup steps, and 0.0006 peak learning rate. These are identical hyperparameters to~\newcite{liu2019b}, except for the number of warmup steps which we changed to 20k for improved training stability. Each model was trained using 64 GPUs for up to 72 hours each. We train three seeds for each data configuration. We validate all models on the public Wiki-103 validation set (see \autoref{app:pretrain}). We use FairSeq~\cite{ott2019fairseq} for the pre-training and fine-tuning experiments.

\begin{table*}[t]
  \centering
  \resizebox{\linewidth}{!}{%
\begin{tabular}{lllllllll}
\toprule
Model & QNLI & RTE & QQP & SST-2 & MRPC & PAWS & MNLI-m/mm & CoLA \\ \midrule
\OR & 92.45 +/- 0.2 & 73.62 +/- 3.1 & 91.25 +/- 0.1 & 93.75 +/- 0.4 & 89.09 +/- 0.9 & 94.49 +/- 0.2 & 86.08 +/- 0.2 / 85.4 +/- 0.2 & 52.45 +/- 21 \\ \midrule
\RIV & 91.65 +/- 0.1 & 70.94 +/- 1.2 & 91.39 +/- 0.1 & 92.46 +/- 0.3 & 86.90 +/- 0.3 & 94.26 +/- 0.2 & 83.79 +/- 0.2 / 83.94 +/- 0.3 & 35.25 +/- 32 \\ 
\RIII & 91.56 +/- 0.4 & 69.75 +/- 2.8 & 91.22 +/- 0.1 & 91.97 +/- 0.5 & 86.22 +/- 0.8 & 94.03 +/- 0.1 & 83.83 +/- 0.2 / 83.71 +/- 0.1 & 40.78 +/- 23 \\
\RII & 90.51 +/- 0.1 & 70.00 +/- 2.5 & 91.33 +/- 0.0 & 91.78 +/- 0.3 & 85.90 +/- 1.2 & 93.53 +/- 0.3 & 83.45 +/- 0.3 / 83.54 +/- 0.3 & 50.83 +/- 5.8 \\
\RI & 89.05 +/- 0.2 & 68.48 +/- 2.5 & 91.01 +/- 0.0 & 90.41 +/- 0.4 & 86.06 +/- 0.8 & 89.69 +/- 0.6 & 82.64 +/- 0.1 / 82.67 +/- 0.2 & 31.08 +/- 10 \\
\midrule
\NP & 77.59 +/- 0.3 & 54.78 +/- 2.2 & 87.78 +/- 0.4 & 83.21 +/- 0.6 & 72.78 +/- 1.6 & 57.22 +/- 1.2 & 63.35 +/- 0.4 / 63.63 +/- 0.2 & 2.37 +/- 3.2 \\
%\RU & 77.69 +/- 0.4 & 53.84 +/- 0.6 & 85.92 +/- 0.1 & 84.00 +/- 0.6 & 71.35 +/- 0.8 & 58.43 +/- 0.3 & 72.10 +/- 0.4 / 72.58 +/- 0.4 & 8.89 +/- 1.40 \\
\RC & 66.94 +/- 9.2 & 53.70 +/- 1.0 & 85.57 +/- 0.1 & 83.17 +/- 1.5 & 70.57 +/- 0.7 & 58.59 +/- 0.3 & 71.93 +/- 0.2 / 71.33 +/- 0.5 & 0.92 +/- 2.1 \\
\RT & 62.17 +/- 0.4 & 52.97 +/- 0.2 & 81.53 +/- 0.2 & 82.0 +/- 0.7 & 70.32 +/- 1.5 & 56.62 +/- 0.0 & 65.70 +/- 0.2 / 65.75 +/- 0.3 & 8.06 +/- 1.6 \\
 \bottomrule
\end{tabular}}
\caption{GLUE and PAWS-Wiki dev set results on different RoBERTa (base) models trained on variants of the BookWiki corpus (with mean and std). The top row is the original model, the middle half contains our primary models under investigation, and the bottom half contains the baselines.}
\label{table:glue_mean}
\end{table*}

\subsection{Fine-tuning tasks}

We evaluate downstream performance using the General Language Understanding and Evaluation (GLUE) benchmark, the Paraphrase Adversaries from Word Scrambling (PAWS) dataset, and various parametric and non-parametric tasks (see \S \ref{sec:probing_results}). 
% Our objective is not to match state-of-the-art in either of these datasets / probes. Rather, we compare the models trained on the original corpus and the randomized corpuses to understand the extent to which these models rely on word order learnt during pre-training.

\xhdr{GLUE} %We use the GLUE \cite{wang2018glue} benchmark tasks to inspect the reliance of word order learning during pre-training. This
The GLUE \cite{wang-etal-2018-glue} benchmark is a collection of 9 datasets for evaluating natural language understanding systems, of which we use Corpus of Linguistic Acceptability \cite[CoLA,][]{cola_warstadt2019neural}, Stanford Sentiment Treebank \cite[SST,][]{sst2_socher2013recursive}, Microsoft Research Paragraph Corpus \cite[MRPC,][]{mrpc_dolan2005automatically}, Quora Question Pairs (QQP)\footnote{\href{http://data.quora.com/First-Quora-Dataset-Release-Question-Pairs}{http://data.quora.com/First-Quora-Dataset-Release-Question-Pairs}}, Multi-Genre NLI \cite[MNLI,][]{williams-etal-2018-broad}, Question NLI \cite[QNLI,][]{rajpurkar-etal-2016-squad, qnli_2_demszky2018transforming}, Recognizing Textual Entailment \cite[RTE,][]{rte1_dagan2005pascal, rte2_haim2006second, rte3_giampiccolo2007third, rte5_bentivogli2009fifth}. 
%DK: I commented this out -
%GLUE is an apt testing ground for our investigation, because it has saturated with respect to ever increasing performance using bigger and bigger Transformer models. 
%Any performance gap between a natural word order model and an unnatural word order model then can be explained solely by pre-trained representation and not by task complexity. 
%Furthermore,
\citet{pham2020} show the word order insensitivity of several GLUE tasks (QQP, SST-2), evaluated on public regularly pre-trained checkpoints.
%\citet{pham2020} show that 75-90\% of the correct BERT predictions remain constant after shuffling, thereby introducing errors 10-25\% of the time. We observe that the gap among original and shuffled pre-training models is actually even smaller.

\xhdr{PAWS} The PAWS task \cite{zhang-etal-2019-paws} consists of predicting whether a given pair of sentences are paraphrases. This dataset contains both paraphrase and non-paraphrase pairs with high lexical overlap, which are generated by controlled word swapping and back translation. Since even a small word swap and perturbation can drastically modify the meaning of the sentence, we hypothesize the randomized pre-trained models will struggle to attain a high performance on PAWS.

\xhdr{Fine-tuning details} We use the same fine-tuning methodology used by \citet{liu2019b}, where we run hyperparameter search over the learning rates $\{1 \times 10^{-5}, 2 \times 10^{-5}, 3 \times 10^{-5}\}$ and batch sizes $\{16, 32\}$ for each model. 
%For pre-training, we train three different runs with three seeds. In our fine-tuning experiments, we select one of these at random, and for our probing experiments (in \autoref{sec:probing_results}) we use all three. 
For the best hyperparam configurations of each model, we fine-tune with 5 different seeds and report the mean and standard deviation for each setting. % (\autoref{table:glue_mean}).
\NP\ is fine-tuned without positional embeddings, matching the way it was pre-trained.

\section{Downstream task results}

In this section, we present the downstream task performance of the models defined in \autoref{sec:experimental_setup}. For evaluation, we report Matthews correlation for CoLA and accuracy for all other tasks. 
%We denote the performance on any task $\mathcal{T}$ to be $\mathcal{A}(\mathcal{T} | D)$, where $\mathcal{T}$ is the task and $D$ is the pre-trained model which is used for fine-tuning.

%\subsection{Impact of word order during pre-training on fine-tuning tasks}

%\subsection{How do different word order permuted pre-training models perform on downstream tasks?}
\subsection{Word order permuted pre-training}
\label{subsec:glue_results}

% Itemized results for reference only, remove in the final draft
% \begin{itemize}
%   \item \textit{Task performance on pre-trained models with scrambled word order is very close to those trained on the original corpus}
%   \item \textit{Task performance monotonically increases with increase in ngrams kept together during pre-training}
%   \item \textit{Large gap between corpus randomization and original model, although much smaller gap between sentence randomization models.}
% \end{itemize}

\begin{figure*}[ht]
    \centering
    \resizebox{\textwidth}{!}{
        \includegraphics{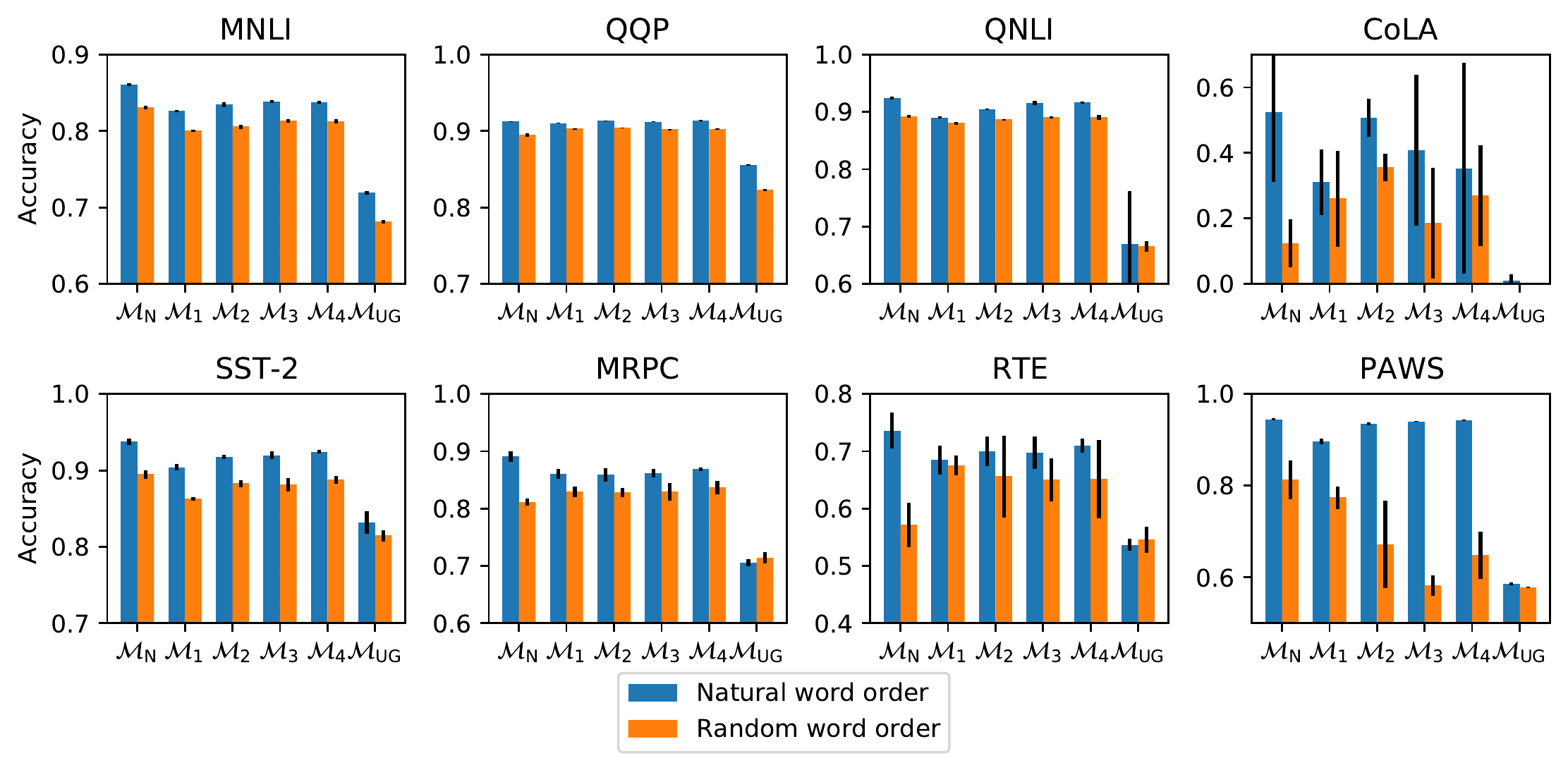}}
    \caption{GLUE \& PAWS task dev performance when finetuned on naturally (blue) and randomly ordered (orange) text, respectively, using pre-trained RoBERTa (base) models trained on different versions of BookWiki corpus.}
    \label{fig:word_order_glue}
\end{figure*}

In our first set of experiments, we finetune the pre-trained models 
%\OR, \RI, \RII, \RIII, \RIV\ and \RC\
on the GLUE and PAWS tasks. 
We report the results in \autoref{table:glue_mean}.\footnote{The \OR\ results are not directly comparable with that of publicly released \texttt{roberta-base} model by \citet{liu2019b}, as that uses the significantly larger 160GB corpus, and is trained for 500K updates. For computational reasons, we restrict our experiments to the 16GB BookWiki corpus and 100K updates, mirroring the RoBERTa ablations.} 
First, we observe that the model without access to distributional or word order information, \RC\ (unigram) performs much worse than \OR\ overall:
\RC\ is $18$ points worse than \OR\ on average across the accuracy-based tasks in \autoref{table:glue_mean} and has essentially no correlation with human judgments on CoLA. \RC\, \NP\ and \RT\ perform comparably on most of the tasks, while achieving surprisingly high scores in QQP and SST-2. However, all three models perform significantly worse on GLUE and PAWS, compared to \OR\ (\autoref{table:glue_mean}, bottom half).
\RC\ reaches up to $71.9$ on MNLI - possibly due to the fact that \RC\ has access to (bags of) words and some phrases (from NER) is beneficial for MNLI. For the majority of tasks, the difference between \NP\ and \RT\ is small - a pure bag of words model performs comparably to a randomly initialized model.

Next, we observe a significant improvement on all tasks when we give models access to sentence-level distributional information during pre-training.
\RI, the model pre-trained on completely shuffled sentences, is on average only $3.3$ points lower than \OR\ on the accuracy-based tasks,
and within $0.3$ points of \OR\ on QQP.
Even on PAWS, which was designed to require knowledge of word order, \RI\ is within $5$ points of \OR.
Randomizing $n$-grams instead of words during pre-training results in a (mostly) smooth increase on these tasks: \RIV, the model pre-trained on shuffled $4$-grams, trails \OR\ by only $1.3$ points on average, and even comes within $0.2$ points of \OR\ on PAWS.
%Compared to \citet{pham2020} results, our observations on the gap between \OR and shuffled pre-train models to be much closer than their observations.
We observe a somewhat different pattern on CoLA, where \RII\ does almost as well as \OR\ and outperforms \RIII\ and \RIV, though we also observe very high variance across random seeds for this task. 
Crucially, we observe that \RI\ outperforms \NP\ by a large margin. This shows that positional embeddings are critical for learning, even when the word orders themselves are not natural.\footnote{Recall, \NP\ is fed natural sentences as \OR\, while not having the ability to learn positional embeddings. To further quantify the effect of positional embeddings, we also investigated the effect of shuffling the entire context window, to keep the co-occurrence information same as \NP\ in \autoref{app:ablations}. We observed this model to be worse than \RI\, but significantly better than \NP\, to support the claim about the importance of positional embeddings while training.}
Overall, these results confirm our hypothesis that RoBERTa's strong performance on downstream tasks can be explained for a large part by the distributional prior.

%\subsection{What contributes to word order learning - pre-training or fine-tuning?}
\subsection{Word order permuted fine-tuning}

There are two possible explanations for the results in \autoref{subsec:glue_results}: either the tasks do not need word order information to be solved, or any necessary word order information can be acquired during fine-tuning. 
To examine this question, we permute the word order during fine-tuning as well.
%A natural question arises following the results from the previous section: does pre-training capture most of the word order information, or does finetuning on the downstream task contribute more to a model's knowledge of word order? To disentangle the cause and effect, we use our set of models to finetune on a randomized version of the downstream tasks. 
%TODO: Do this at the sentence level or be MUCH more explicit that this is NOT at the sentence level:
Concretely, for each task, we construct a unigram order-randomized version of each example in the fine-tuning training set using $\mathcal{F}_1$. We then fine-tune our pre-trained models on this shuffled data and evaluate task performance. For all experiments, we evaluate and perform early stopping on the original, natural word order dev set, in order to conduct a fair evaluation on the exact same optimization setup for all models. 
% The hypothesis is that if the source of word order is in the pre-trained corpus, then $\mathcal{A}(\mathcal{T} | D) - \mathcal{A}(\mathcal{T}_{\mathcal{F}_1} | D) > 0$. 
%The relative magnitude of $\delta$, $\Delta_{D} = \frac{\delta}{\mathcal{A}(\mathcal{T} | D)}$, should provide an indication that which tasks rely on word order information from the source corpus.

% \begin{itemize}
%     \item \textit{Models learn word order from downstream tasks than from the pre-trained checkpoints}
%     \item \textit{Some tasks are more reliant on word order information than the rest (PAWS, CoLA, RTE)} (Similar finding as \cite{pham2020})
%     \item \textit{For some tasks, model learn random word order better with the help of random pre-training (RTE, CoLA, MRPC)}
% \end{itemize}

% If the task in question is the source of word order for these models, then the downstream performance on the unnatural corpus should be worse than that on the natural corpus. Specifically, we take the same tasks we consider above (GLUE and PAWS) and construct a randomized training set using our permutation function $\mathcal{F}$. 

% Self note: maybe the plot is not correct? Or maybe need to think of a metric

Our results in \autoref{fig:word_order_glue} provide some evidence for both hypotheses.
%provide some evidence for both hypotheses, but primarily show that shuffled models \emph{are} learning to use word order-related information during fine-tuning.
On QQP and QNLI, accuracy decreases only slightly for models fine-tuned on shuffled data. %, suggesting that word order is not very important for these tasks.
Models can also achieve above $80\%$ accuracy on MNLI, SST-2, and MRPC when fine-tuned on shuffled data,
suggesting that purely lexical information is quite useful on its own.\footnote{
This finding is compatible with the observation of \citet{gupta-etal-2021-bert} and \citet{sinha2020b} who train on a randomized training corpus for MRPC, QQP, SST-2 and MNLI.} 

%Yet, we observe high task performance ($\ge  80\%$) for MNLI, QQP, SST-2, MRPC, and QNLI when the underlying pre-trained model is \OR, \RI, \RII, \RIII, \RIV. 
%This finding suggests that irrespective of the prior, models are able to learn specific lexical information from the training distribution (similar to word2vec), enabling them to provide competitive advantage on the fine-tuning task. %in spite of the data being out-of-domain for the model from a syntax perspective. 

On the other hand, for all datasets besides QQP and QNLI, we see noticeable drops in accuracy when fine-tuning on shuffled data and testing on normal order, both for \OR\ and for shuffled models \RI\ through \RIV.
This suggests both that word order information is useful for these tasks, and that shuffled models must be learning to use word order information during fine-tuning.\footnote{We perform additional experiments on how the model representations change during fine-tuning for shuffled training using Risannen Data Analysis in \autoref{app:rda_analysis}.}
Having word order during fine-tuning is especially important for achieving high accuracy on CoLA, RTE (cf. \citealt{pham2020}), as well as PAWS, 
suggesting that these tasks are the most word order reliant. Recent research \cite{yu-ettinger-2021-interplay} raised some questions about potential artefacts inflating performance on PAWS: their swapping-distance cue of appears consistent both with our finding of high PAWS performance for n-gram shuffled models in \autoref{table:glue_mean}, and with our PAWS results in \autoref{fig:word_order_glue}, which suggests that PAWS performance does in fact rely to some extent on natural word order  at the fine-tuning stage.
%\citet{pham2020} similarly found word order to be most important for CoLA and RTE out of all GLUE tasks.

%On MNLI, SST-2, and MRPC, we see larger drops in accuracy, suggesting that modeling word order-dependent cues is important for these tasks. 
%Nonetheless, models can reach above $80\%$ accuracy on these datasets, indicating that purely lexical information is quite useful on its own.
%However, for other tasks like PAWS and CoLA, fine-tuning on shuffled task data results in large decreases in test accuracy.
%When we train on word order randomized versions of the tasks evaluate on the original dev sets, the setup is \textit{out-of-domain} for the model leading to the expectation of a low task performance. 
%Yet, we observe high task performance ($\ge  80\%$) for MNLI, QQP, SST-2, MRPC, and QNLI when the underlying pre-trained model is \OR, \RI, \RII, \RIII, \RIV. 
%This finding suggests that irrespective of the prior, models are able to learn specific lexical information from the training distribution (similar to word2vec), enabling them to provide competitive advantage on the fine-tuning task. %in spite of the data being out-of-domain for the model from a syntax perspective. 

Finally, for CoLA, MRPC, and RTE, performance is higher after fine-tuning on shuffled data for \RI\ than \OR.
We hypothesize that \OR\ represents shuffled and non-shuffled sentences very differently, resulting in a domain mismatch problem when fine-tuning on shuffled data but evaluating on non-shuffled data.\footnote{We further study the domain mismatch problem by evaluating on shuffled data \textit{after} fine-tuning on the shuffled data for models in \autoref{app:rand_eval}. We observe that models improves their scores on evaluation on shuffled data when the training data source is changed from natural to shuffled - highlighting domain match effect.} Since \RI\ never learns to be sensitive to word order during pre-training or fine-tuning, it does not suffer from that issue. Our results in this section also highlights the issues with these datasets, concurrent to the findings that many GLUE tasks does not need sophisticated linguistic knowledge to solve, as models typically tend to exploit the statistical artefacts and spurious correlations during fine-tuning (cf. \citealt{gururangan-etal-2018-annotation, poliak-etal-2018-hypothesis,tsuchiya-2018-performance,mccoy-etal-2019-right}). However, our results overwhelmingly support the fact that word order does not matter during pre-training, if the model has the opportunity to learn the necessary information about word order during fine-tuning.

% TODO: cite mccoy2019, cite prasanna's paper

%the task performance on word order randomized data increases when the pre-trained model is \RI instead of \OR. For these tasks, training on randomized sentences having a randomized prior (Recall, \RI = $ \mathcal{F}_1(\textrm{BookWiki})$) creates an \textit{in-domain} advantage in downstream performance. 
%The distributional prior of unnatural sentences being beneficial for the unnatural versions of these tasks leads us to conclude that they are \textit{word order reliant}.
%These results align with \citet{pham2020}, although we always evaluate on the natural word order sentences, thus yielding stronger empirical evidence that some tasks rely more on their respective training corpus to learn the correct word order.
%These results also indicate the capacity of $OR$ to distinguish more between normal and shuffled sentences, which results in big domain shift between finetuning on shuffled and testing on normal sentences.
% mention gupta et al

% Given the relatively small size of the training datasets of these tasks, it can be concluded that smaller downstream tasks are able to provide better opportunity of 

% how do I explain this?

% Since all models are evaluated on natural word order corpus (dev set of individual datasets), this conclusively proves that word order information crucial to perform those tasks are learned by the models during fine-tuning more than that of pre-training.

\section{Probing results}\label{sec:probing_results}

%Downstream tasks paint only a partial picture of the information contained in the pre-trained representation.
To investigate how much syntactic information is contained in the MLM representations, we evaluate several probing tasks on our trained models. We consider two classes of probes: \textit{parametric} probes, which make use of learnable parameters, and \textit{non-parametric} probes, which directly examine the language model's predictions.

\subsection{Parametric Probing}
\label{subsec:param_probing}

To probe our models for syntactic, semantic and other linguistic properties, we investigate dependency parsing using Pareto probing~\cite{pimentel-etal-2020-pareto} and the probing tasks from~\newcite{conneau-etal-2018-cram} in SentEval~\cite{conneau-kiela-2018-senteval}.

\subsubsection{Syntactic Probing}

\citet{pimentel-etal-2020-pareto} proposed a framework based on Pareto optimality to probe for syntactic information in contextual representations. They suggest that an optimal probe should balance optimal performance on the probing task with the complexity of the probe. Following their setup, we use the ``difficult'' probe: dependency parsing (DEP). We also investigate the ``easy'' probes, dependency arc labeling (DAL) and POS tag prediction (POS), results are reported in Appendix \ref{app:pareto_probes}.
We probe with Linear and MLP probes, and inspect the task accuracy in terms of Unlabeled Attachment Score (UAS).
The dependency parsing probe used in \citet{pimentel-etal-2020-pareto} builds on the Biaffine Dependency Parser~\citep{dozat2016deep}, but with simple MLPs on top of the Transformer representations.%
%the complexity of the probe reduced by using only simple MLPs on top of the Transformer representations.
%removing the LSTM layer to restrict access to context. This probe consists of two identical MLPs - one to process the heads of the dependencies, and another to process the tails. The final biaffine transformation is simply a learned mapping function among output of the MLPs which is normalized to obtain the probabilities.
\footnote{We experimented with a much stronger, state-of-the-art Second order Tree CRF Neural Dependency Parser \cite{zhang-etal-2020-efficient}, but did not observe any difference in UAS with different pre-trained models (see \autoref{app:sota_dep})}

% If we need the equation, uncomment this block:
% \begin{equation}
% \begin{split}
%     h_{i, \textrm{head}} = \textrm{MLP}_{\textrm{head}}(h_i) \\
%     h_{i, \textrm{tail}} = \textrm{MLP}_{\textrm{tail}}(h_i) \\
%     l_{i,j} = h_{i,\textrm{head}} \dot W \dot h_{j, \textrm{tail}} \\
%     p_{\textrm{parse}}(\textrm{head} = i | \textrm{tail} = j) = \frac{e^{l_{i,j}}}{\sum_{\hat{i}} e^{l_{\hat{i}},j}}
% \end{split}
% \end{equation}

% Concretely, the paper suggests to compute accuracy on the Dependency parsing task on a set of hyperparameters, and for the same set of hyperparameters evaluate memorization capacity on a label-shuffled corpus. In the end, we obtain a ``hypervolume'' which is closer to an Area under the curve (AUC) score of accuracy vs memorization. Contextualized representations are then compared against this hypervolume. As recommended in the paper, we look at Dependency Parsing, and we also observe the maximum accuracy (UAS score) on the test set for different classes of probes. If the contextualized representations in RoBERTa rely on natural word order, then we should expect large gap between original model and word order scrambled models.

% \begin{itemize}
%     \item \textit{Dependency parsing using Linear and MLP probes offer more gap than downstream tasks.}
%     \item \textit{Training on larger amount of data narrows the gap between UAS scores. (EWT having 12,543 sentences compared to PTB having 39,832 sentences. }
% \end{itemize}

\xhdr{Training setup} Similar to the setup by \citet{pimentel-etal-2020-pareto}, we run 50 random hyperparameter searches on both MLP and Linear probes by uniformly sampling from the number of layers (0-5), dropout (0-0.5), log-uniform hidden size $[2^{5}, 2^{10}]$. We triple this experiment size by evaluating on three pre-trained models of different seeds for each model configuration. 
We consider \citeauthor{pimentel-etal-2020-pareto}'s English dataset, derived from Universal Dependencies EWT (UD EWT) \cite{bies2012english, silveira2014gold} which contains 12,543 training sentences. 
%Since \citet{pimentel-etal-2020-pareto} explored multi-lingual corpora, they only experimented with a single English dataset, derived from Universal Dependencies EWT (UD EWT) \cite{bies2012english, silveira2014gold} containing 12,543 training sentences. 
Additionally, we experiment on the Penn Treebank dataset (PTB), which contains 39,832 training sentences.\footnote{PTB data \citep{kitaev-etal-2019-multilingual} is used from \href{https://github.com/nikitakit/self-attentive-parser/tree/master/data}{github.com/nikitakit/self-attentive-parser/tree/master/data}.}
%Since \citet{pimentel-etal-2020-pareto} explored multi-lingual corpora, they only experimented with a single English dataset, derived from Universal Dependencies EWT (UD EWT) \cite{bies2012english, silveira2014gold} containing 12,543 training sentences. Additionally, we experiment on the Penn Treebank dataset (PTB), which contains 39,832 training sentences.\footnote{PTB data \citep{kitaev-etal-2019-multilingual} is used from \href{https://github.com/nikitakit/self-attentive-parser/tree/master/data}{github.com/nikitakit/self-attentive-parser/tree/master/data}.}
We report the mean test accuracy over three seeds for the best dev set accuracy for each task.\footnote{\citet{pimentel-etal-2020-pareto} propose computing the \textit{Pareto Hypervolume} over all hyperparameters in each task. We did not observe a significant difference in the hypervolumes for the models, as reported in \autoref{app:pareto_probes}.}

\begin{table}[]
\centering
\resizebox{\linewidth}{!}{%
\begin{tabular}{l|rl|rl}
\toprule
Model & \multicolumn{2}{c|}{UD EWT} & \multicolumn{2}{c}{PTB} \\ \hline
 & MLP & Linear & MLP & Linear \\ \cline{2-5} 
\OR & 80.41 +/- 0.85 & 66.26 +/- 1.59 & 86.99 +/- 1.49 & 66.47 +/- 2.77 \\\midrule
\RIV & 78.04 +/- 2.06 & 65.61 +/- 1.99 & 85.62 +/- 1.09 & 66.49 +/- 2.02 \\
\RIII & 77.80 +/- 3.09 & 64.89 +/- 2.63 & 85.89 +/- 1.01 & 66.11 +/- 1.68 \\
\RII & 78.22 +/- 0.88 & 64.96 +/- 2.32 & 84.72 +/- 0.55 & 64.69 +/- 2.50 \\
\RI & 69.26 +/- 6.00 & 56.24 +/- 5.05 & 79.43 +/- 0.96 & 57.20 +/- 2.76 \\\midrule
\RC & 74.15 +/- 0.93 & 65.69 +/- 7.35 & 80.07 +/- 0.79 & 57.28 +/- 1.42 \\
\bottomrule
\end{tabular}%
}
\caption{Unlabeled Attachment Score (UAS) (mean and std) on the dependency parsing task (DEP) on two datasets, UD EWT and PTB, using the Pareto Probing framework \cite{pimentel-etal-2020-pareto}.}
\label{tab:pareto_dependency}
\end{table}

\begin{table*}[ht]
  \centering
  \resizebox{\linewidth}{!}{%
\begin{tabular}{lrrrrrrrrrr}
\toprule
                      \textbf{Model} &          \textbf{Length} &     \textbf{WordContent} &           \textbf{TreeDepth} & \textbf{TopConstituents} &     \textbf{BigramShift} &           \textbf{Tense} & \textbf{SubjNumber} &  \textbf{ObjNumber} &       \textbf{OddManOut} & \textbf{CoordInversion} \\
                      & (Surface) & (Surface) & (Syntactic) & (Syntactic)  & (Syntactic) & (Semantic)  & (Semantic) & (Semantic) & (Semantic) & (Semantic) \\
\midrule
\OR &  78.92 +/- 1.91 &  31.83 +/- 1.75 &  35.97 +/- 1.38 &  \textbf{78.26} +/- 4.08 &  \textbf{81.82} +/- 0.55 &  87.83 +/- 0.51 &  85.05 +/- 1.23 &  75.94 +/- 0.68 &   58.40 +/- 0.33 &        \textbf{70.87} +/- 2.46 \\\midrule
\RIV &  92.88 +/- 0.15 &  57.78 +/- 0.36 &  40.05 +/- 0.29 &   72.50 +/- 0.51 &  76.12 +/- 0.29 &  88.32 +/- 0.13 &  \textbf{85.65} +/- 0.13 &  82.95 +/- 0.05 &   \textbf{58.89} +/- 0.30 &        61.31 +/- 0.19 \\
\RIII &  91.52 +/- 0.16 &  48.81 +/- 0.26 &  38.63 +/- 0.61 &  70.29 +/- 0.31 &  77.36 +/- 0.12 &  86.74 +/- 0.12 &  83.83 +/- 0.38 &  80.99 +/- 0.26 &  57.01 +/- 0.21 &         60.00 +/- 0.26 \\
\RII &  \textbf{93.54} +/- 0.29 &  62.52 +/- 0.21 &   \textbf{41.40} +/- 0.32 &  74.31 +/- 0.29 &  75.44 +/- 0.14 &  \textbf{87.91} +/- 0.35 &  84.88 +/- 0.11 &  83.98 +/- 0.14 &   57.60 +/- 0.36 &        59.46 +/- 0.37 \\
\RI &  88.33 +/- 0.14 &  \textbf{64.03} +/- 0.34 &   40.24 +/- 0.20 &  70.94 +/- 0.38 &   58.37 +/- 0.40 &  87.88 +/- 0.08 &  83.49 +/- 0.12 &  \textbf{83.44} +/- 0.06 &  56.51 +/- 0.26 &         56.98 +/- 0.50 \\
\midrule
\RC &  86.69 +/- 0.33 &   36.60 +/- 0.33 &  32.53 +/- 0.76 &   61.54 +/- 0.60 &  57.42 +/- 0.04 &  68.45 +/- 0.23 &  71.25 +/- 0.12 &  66.63 +/- 0.21 &   50.06 +/- 0.40 &        56.26 +/- 0.17 \\ 
%\hline
%\texttt{RB} &   75.86 &    27.4 &   34.02  &   71.85  &   88.43 &   88.32  &   84.01 &   81.83 &   66.41  &         70.68 \\
\bottomrule
\end{tabular}}
\caption{SentEval Probing \cite{conneau-etal-2018-cram, conneau-kiela-2018-senteval} results (with mean and std) on different model variants.} 
%Results on the publicly available Roberta (base) (\texttt{RB}) from \citet{liu2019b} provided for comparison.
\label{table:senteval}
\end{table*}

\xhdr{Results} We observe that 
the UAS scores follow a similar linear trend as the fine-tuning results in that \RI $\approx$ \RC < \RII < \RIII < \RIV < \OR\ (\autoref{tab:pareto_dependency}). 
Surprisingly, \RC\ probing scores seem to be somewhat better than \RI\ (though with large overlap in their standard deviations), even though \RC\ cannot learn information related to either word order or co-occurrence patterns. 
The performance gap appears to be task- and probe specific.
%The gap in UAS between \RI and \OR\ for UD EWT is higher than that of the counterpart in PTB ($11.15$ vs. $7.56$). This trend is also mildly reflected in the case of the Linear probe ($10.02$ vs. $9.27$ UD EWT). 
We observe a low performance gap in several scenarios, the lowest being between \OR\ vs. \RIII/\RIV, for PTB using the both MLP and Linear probes. 
%These results are an indicator of possible deficiencies in the method of testing representations for syntactic information.

% For example, for POS Tagging experiments using PTB and using MLP probe, $OR$ achieves 97.07\% vs $R_1$ which achieves 95.33\% accuracy on test set.
%Our findings corroborate with that of \citet{pimentel-etal-2020-pareto}, in that harder tasks are better to understand the power of test representation. 

% is an indicator of possible deficiencies in the method of testing representations for syntactic information. 
% The low probe UAS and equivalently high downstream task accuracy for $R_1$ strongly suggests that ``learning the classical NLP pipeline'' is probably not a big factor in BERT's success.

%We conduct further experiments using Pareto Probing framework \cite{pimentel-etal-2020-pareto} on other, easier probing tasks. We present the results of probing on POS tagging task in \autoref{tab:pareto_pos_tag} and Dependency arc labelling task in \autoref{tab:pareto_dep_label}. Since both of these tasks are simpler than dependency parsing, the gap between the $OR$ and unnaturally pre-trained models reduces even more drastically. For example, for POS Tagging experiments using PTB and using MLP probe, $OR$ achieves 97.07\% vs $R_1$ which achieves 95.33\% accuracy on test set.

\subsubsection{SentEval Probes}

We also investigate the suite of 10 probing tasks~\cite{conneau-etal-2018-cram} available in the SentEval toolkit~\cite{conneau-kiela-2018-senteval}. This suite contains a range of semantic, syntactic and surface level tasks. \citet{jawahar2019a} utilize this set of probing tasks to arrive at the conclusion that %that BERT-based models naturally learn syntax -
``\textit{BERT embeds a rich hierarchy of linguistic signals: surface information at the bottom, syntactic information in the middle, semantic information at the top}''. We re-examine this hypothesis by using the same probing method and comparing against models trained with random word order.

\xhdr{Training setup} We run the probes on the final layer of each of our pre-trained models for three seeds, while keeping the encoder frozen. % It is important to note that SentEval probe evaluates on frozen encoder representations. 
SentEval %obtains the sentence representations from the model under investigation, and then 
trains %linear and two-layer MLP 
probes on top of fixed representations individually for each task. We follow the recommended setup and run grid search over the following hyperparams: number of hidden layer dimensions ($[0,50,100,200]$), dropout ($[0, 0.1, 0.2]$), 4 epochs, 64 batch size. 
%For all hyperparameters, 
We select the best performance based on the dev set, and report the test set accuracy.

\xhdr{Results} We provide the results in \autoref{table:senteval}. The \OR\ pre-trained model scores better than the unnatural word order models for only one out of five semantic tasks and in none of the lexical tasks. However, \OR\ does score higher for two out of three syntactic tasks. Even for these two syntactic tasks, the gap among \RC\ and \OR\ is much higher than \RI\ and \OR. These results show that while natural word order is useful for at least some probing tasks, the distributional prior of randomized models alone is enough to achieve a reasonably high accuracy on syntax sensitive probing.
%These dismal numbers casts a grim outlook on the ``discovering NLP pipeline" aspect of BERT-based models. However, it could also be the case that the probes can be solved by leveraging tail distributions [need more info/citations].
% KS: Douwe, can you add one-liner on why the probes are not performing?

% \begin{itemize}
%     \item \textit{SentEval, consisting of various surface, syntactic and semantic probes, fail to discern properly among randomized checkpoints. }
%     \item \textit{Only BiGramShift, TopConstituents and CoordinationInversion has better scores than the original model.}
% \end{itemize}

\subsection{Non-Parametric Probing}
\label{sec:non_param_probe}

How to probe effectively with parametric probes is a matter of much recent debate~\citep{hall-maudslay-etal-2020-tale, belinkov2021probing}. From our results so far, it is unclear whether parametric probing meaningfully distinguishes models trained with corrupted word order from those trained with normal orders. Thus, we also investigate non-parametric probes~\cite{linzen-etal-2016-assessing,marvin-linzen-2018-targeted,gulordava2018} using the formulation of \citet{goldberga} and \citet{wolf2019}.% Since these probes do not contain any learnable parameters, they are called ``non-parametric''. 

We consider a set of non-parametric probes that use a range of sentences varying in their linguistic properties. For each, the objective is for a pre-trained model to provide higher probability to a grammatically correct word than to an incorrect one. 
Since both the correct and incorrect options occupy the same sentential position, we call them ``focus words''. \citet{linzen-etal-2016-assessing} use sentences from Wikipedia containing present-tense verbs, and compare the probability assigned by the encoder to plural vs. singular forms of the verb; they focus on sentences containing at least one noun between the verb and its subject, known as ``agreement attractors.'' \citet{gulordava2018} instead replace focus words with random substitutes from the same part-of-speech and inflection. Finally, \citet{marvin-linzen-2018-targeted} construct minimal pairs of grammatical and ungrammatical sentences, and compare the model's probability for the words that differ.

\xhdr{Setup} In our experiments, we mask the focus words in the stimuli and compute the probability of the correct and incorrect token respectively.
To handle Byte-Pair Encoding (BPE), we use the WordPiece \cite{wu2016googles} tokens prepended with a space. %, as used in RoBERTa to compute the probability of a correct focus word and an incorrect focus word. D: this says the same thing as the first clause of this sentence, removed it
We observe that the \citet{linzen-etal-2016-assessing} and \citet{gulordava2018} datasets are skewed towards singular focus words, which could disproportionately help weaker models that just happen to assign more probability mass to singular focus words. To counter this, we balance these datasets to have an equal number of singular and plural focus words by upsampling, and report the aggregated and balanced results in \autoref{tab:non_param_comb} (see \autoref{app:non_par_probes} for more detailed results). We verify our experiments by using three pre-trained models with different seeds for each model configuration.

\xhdr{Results}
We observe for the \citet{linzen-etal-2016-assessing} and \citet{marvin-linzen-2018-targeted} datasets that the gap between the \OR\ and randomization models is relatively large.
The \citet{gulordava2018} dataset shows a smaller gap between \OR\ and the randomization models. While some randomization models (e.g., \RII, \RIII, and \RIV) performed quite similarly to \OR\ according to the parametric probes, they all are markedly worse than \OR\ according to the non-parametric ones. This suggests that non-parametric probes identify certain syntax-related modeling failures
that parametric ones do not.

% \begin{table}
%   \centering
%   \resizebox{\linewidth}{!}{%
% \begin{tabular}{llll}
% Model & \citet{linzen-etal-2016-assessing} & \citet{gulordava2018} & \citet{marvin2018a} \\
% \midrule
% \OR & 91.17 (2.61) [23.26] & 68.66 (11.56) [1.97] & 88.21 (6.71) [14.53]\\
% \RC & 65.36 (7.06) [1e-4] & 60.88 (24.26) [1e-4] & 50.13 (0.24) [1e-4]\\
% \RI & 58.96 (1.82) [0.07] & 68.1 (14.42) [0.03] & 70.45 (11.42) [0.49]\\
% \RII & 61.27 (3.07) [1.01] & 60.2 (7.64) [1e-4] & 74.08 (14.33) [1.06]\\
% \RIII & 64.6 (2.71)  [2.25] & 66.1 (5.99) [0.24] & 73.78 (15.64) [1.62]\\
% \RIV & 66.93 (3.2)  [3.22] & 69.47 (4.99) [0.46] & 70.8 (12.6) [3.84]\\
% \bottomrule
% \end{tabular}}
% \caption{Mean accuracy of non-parametric probing on different stimuli datasets. Values in paranthesis reflects the standard deviation over sub-tasks. Values in square brackets indicate the mean probability difference.}
% \label{tab:non_param_comb}
% \end{table}

\begin{table}
  \centering
  \resizebox{\linewidth}{!}{%
\begin{tabular}{lrrr}
\toprule
Model & \citet{linzen-etal-2016-assessing} $^*$ & \citet{gulordava2018} $^*$ & \citet{marvin-linzen-2018-targeted} \\
\midrule
\OR & 91.17 +/- 2.6 & 68.66 +/- 11.6 & 88.05 +/- 6.5 \\
\RIV & 66.93 +/- 3.2 & 69.47 +/- 4.9  & 70.66 +/- 12.5 \\
\RIII & 64.60 +/- 2.7 & 66.10 +/- 5.9  & 73.82 +/- 15.7 \\
\RII & 61.27 +/- 3.1 & 60.20 +/- 7.6  & 73.95 +/- 14.3 \\
\RI & 58.96 +/- 1.8 & 68.10 +/- 14.4 & 70.69 +/- 11.6 \\
\RC & 65.36 +/- 7.1 & 60.88 +/- 24.3 & 50.10 +/- 0.2 \\
\bottomrule
\end{tabular}}
\caption{Mean (and std) non-parametric probing accuracy on different datasets. $^*$ indicates rebalanced datasets, see \autoref{app:non_par_probes} for more details.}
\label{tab:non_param_comb}
\end{table}

%We recommend that future work consider stimuli-based probing in greater detail to uncover the syntax information in contextual representations.

% Moving analysis to appendix
%\section{Analysis}

%In this section, we perform further analysis to determine how much word order information is present in the RoBERTa representation after pre-training. Specifically, we first use perplexity as a proxy for measuring whether certain word order information is contained in the representation after pre-training. Secondly, we investigate the effect of the learned word order during the early stages of fine-tuning.

%\subsection{Does pre-training with random word order detect n-gram randomization?}

\section{Discussion}

The assumption that word order information is crucial for any classical NLP pipeline (especially for English) is deeply ingrained in our understanding of syntax itself \citep{chomsky1957syntactic}: without order, many linguistic constructs are undefined. % (e.g.\ dependency or constituency parses would no longer be syntactic trees, what would sentences be but mere lists of words).
Our fine-tuning results in \autoref{subsec:glue_results} and parametric probing results in \autoref{subsec:param_probing}, however, suggests that MLMs do not need to rely much on word order to achieve high accuracy, bringing into question previous claims that they learn a ``classical NLP pipeline.''

% The fine-tuning results in \autoref{subsec:glue_results} and the parametric probing results in \autoref{subsec:param_probing} suggest that MLM does not need to rely on the ``classical NLP pipeline'' for achieving high accuracy, assuming that such a pipeline would rely upon word order. 
% The assumption that word order information is crucial for any classical NLP pipeline (especially for English) is deeply ingrained in our understanding of syntax itself \citep{chomsky1957syntactic}: without order, most linguistic constructs are undefined (e.g.\ dependency or constituency parses would no longer be syntactic trees, what would sentences be but mere lists of words).

%One might argue that word order may just not be as important for syntactic understanding of English as we had previously thought, but 
%We argue that word order information is crucial 
%, because 
%Our results indicate that the notion of ``syntax tree'' extracted by MLM-style pre-trained models is no longer a tree. Rather, the models learn to extract a distributional, fuzzy representation of syntax trees.
%
One might ask, though, whether an NLP pipeline would really need natural word order at all: can transformers not simply learn what the correct word order is from unordered text?
%
%One interpretation of our results could be that BERT can figure out the correct word order for itself during pre-training. However, 
First, the lower non-parametric probing accuracies of the randomized models indicate that they are not able to accurately reconstruct the original word order (see also \autoref{app:ablations}).
But even if models were able to ``unshuffle'' the words under our unnatural pre-training set up, they would only be doing so based on distributional information. 
Models would then abductively learn only the most likely word order. %---as evidenced by the effectiveness of the distributional prior alone in achieving good enough fine-tuning and probing performance, high perplexity for randomized models on the original data, and our non-parametric probing results. % showing that these models cannot recover word order nearly as well. 
While models might infer a distribution over possible orders and use that information to structure their representations \citep{papadimitriou-etal-2021-deep}, syntax is not about \emph{possible} or even  \emph{the most likely} orders: it is about the \emph{actual} order. That is, even if one concludes in the end that Transformers are able to perform word order reconstruction based on distributional information, and recover almost all downstream performance based solely on that, we ought to be a lot more careful when making claims about what our evaluation datasets are telling us.

% IS ENGLISH MORE ORDER-FREE THAN WE THOUGHT?

% NONPARAMETRIC PROBES SHOW THAT IT CLEARLY CANNOT RECOVER THE WORD ORDER AS WELL AS THE OR MODEL

% LOCALITY VS DISTRIBUTIONAL INFO

Thus, our results seem to suggest that we may need to revisit what we mean by ``linguistic structure,'' and perhaps subsequently acknowledge that we may not need human-like linguistic abilities for most NLP tasks. Or, our results can be interpreted as evidence that we need to develop more challenging and more comprehensive evaluations%
%(cf. \citealt{bowman2021will})
, if we genuinely want to measure linguistic abilities, however those are defined, in NLP models.

% Todo: talk about locality information

% TODO: add a limitation section here

 %If we decide not to blame the inadequacies of our current evaluation paradigm, one alternative interpretation of our results is that English is somehow less dependent on word order than expected, perhaps with distributional information playing a much bigger role than syntactic structures. 

% Notes from discussion with Douwe

% Q: is it obvious that removing ordering information from the input prevents the model from learning any deeper linguistic structure? Could another interpretation of your results be that the model is learning to figure out the correct word order for itself during pre-training?

% A: even if it has learned to reorder, it can only do so based on distributional information, meaning it learns the distributionally most likely order and ignores actual order -- and as it turns out, that gets you almost all the way there.

% Q: the underlying question in the above is: do you actually need order for "deeper linguistic structure" / the classical NLP pipeline / whatever?

% A: Order is required for "linguistic structure" because otherwise most linguistic information is undefined : subject verb order in English for example; dependency tree is undefined; constituent parses are undefined.. etc etc. Basically, our syntax tree which we thought is so important is no longer a tree

% So this suggests: we need harder evaluations. Or we need to revise our meaning of "linguistic structure".

There are many interesting and potentially exciting avenues for future work that we could not explore due to limitation of space. An interesting question revolves around whether this phenomenon is more pronounced for English than for other languages. It is natural to wonder whether more word-order flexible or morphologically-rich languages would suffer from the same problem. Using the methods discussed in this work, we could imagine devising a way to determine the degree of order-dependence for tasks across languages. Another possible extension pertains to other tasks, including extractive question answering (QA) or sequence tagging, for which we can also to determine whether word order information is acquired downstream or during pre-training. 

The sensitivity of generative models to word order permuted input could also be investigated further. %, as several recent works highlights the fact that NLU models and tasks \cite{sinha2020b, pham2020, gupta-etal-2021-bert} are mostly word order in-sensitive. 
Recent work by \citet{parthasarathi2021want} begins this discussion, by showing that a Machine Translation (MT) model can often arrive at the gold source translation when provided with input sentences that have had their words permuted using parse trees. Relatedly, \newcite{alleman-etal-2021-syntactic} also investigates targeted parse-tree-based perturbations as a means of evaluating model robustness. \newcite{oconnor2021context} also demonstrate the insensitivity of Transformers towards syntax manipulations while achieving low perplexity in language modeling tasks. Exploring model sensitivity to word order permutations for approaches that unify generation and classification (e.g., multitasking) could also be interesting future work. 

\section{Conclusion}

In this work, we revisited the hypothesis that masked language modelling's impressive performance can be explained in part by its ability to learn classical NLP pipelines. We investigated targeted pre-training on sentences with various degrees of randomization in their word order, and observed overwhelmingly that MLM's success is most likely not due to its ability to discover syntactic and semantic mechanisms necessary for a traditional language processing pipeline during pre-training. Instead, our experiments suggest that MLM's success can largely be explained by it having learned higher-order distributional statistics that make for a useful prior for subsequent fine-tuning. These results should hopefully encourage the development of better, more challenging tasks that require sophisticated reasoning, and harder probes to narrow down what exact linguistic information is present in the representations learned by our models.
% Adding one line in place of removed section of references
%(DK: not a fan of this as the last sentence though)
%Our results can also lead to investigating synthetic sources  \cite{kataoka2021, papadimitriou2020} as an alternative to leveraging massive amount of data during pre-training.

\section*{Acknowledgements}

We thank Tiago Pimentel, Shruti Bhosale, Naman Goyal, Shagun Sodhani, Sylke Gosen, Prasanna Parasarathi, Kyunghyun Cho, Mona Diab, Brenden Lake, Myle Ott, Ethan Perez, and Mike Lewis for their help in resolving technical doubts during experimentation and/or feedback on an earlier draft. We also thank the anonymous reviewers for their constructive feedback during the reviewing phase, which helped polish the paper to its current state.

% Entries for the entire Anthology, followed by custom entries
\bibliography{anthology,custom}
\bibliographystyle{acl_natbib}

\clearpage
\appendix

% Someone please double-triple-check that this isn't ridiculous:

\section{From Word2vec to BERT in 4 steps}
\label{app:bert-is-word2vec}

Take the basic parameterization of skipgram word2vec \cite{mikolov2013}:

 \begin{equation}
     p(t\;|\;w;\theta) = \frac{e^{f(t, w)}}{\sum_{t'\in V} e ^{f(t', w)}}
 \end{equation}
 
where $t$ is the target, $w$ is a word in the context, $V$ is the set of all possible context words and $f$ is simply the dot product.

In actual word2vec, we would use negative sampling within a given window size and optimize $\log \sigma(w \cdot t) + k \cdot \mathbb{E}_{t' \in P} \log \sigma (-w \cdot t')$ computed over context $C(w) = \{w_{i-k}, ..., w_{i-1}, w_{i+1}, w_{i+k}\}$ for word index $i$, window size $2k$ and unigram probability distribution $P$. It has been shown that optimizing this objective is close to learning the shifted PPMI distribution \cite{levy2015improving}.

\paragraph{Step 1: BPE} One reason for not computing the full softmax is that it becomes a prohibitively expensive matrix multiplication with large vocabulary $V$. A solution is to tokenize based on subword units, e.g. BPE, to ensure a smaller total vocabulary $U$ in the softmax denominator. Doing so makes the matrix multiplication feasible, at least on GPU. It also ensures we have sufficient coverage over the words in our vocabulary.

\paragraph{Step 2: Defenestration} Next, replace the local context window with the entire sentence, while masking out the target word, i.e., $C(t) = \{w \in S\ : w \neq t\}$ where $S$ is the sentence containing $w$.

\paragraph{Step 3: Non-linearity} Replace the pairwise word-level dot product $f(w, t)$ with a fancy non-linear function, say a sequence of multi-head self attention layers, $g(t, C(t))$, that takes as input the entire sentence-with-mask, and you get:

 \begin{align*}
     p(t\;|\;C(t);\theta) = \frac{e^{g(t, C(t))}}{\sum_{t'\in U} e ^{g(t', C(t))}}
 \end{align*}
 
\paragraph{Step 4: Sprinkle data and compute} You have BERT. Now all you need is enough data and compute, and perhaps some optimization tricks. Make sure to update the parameters in your model $g$ when fine-tuning, rather than keeping them fixed, for optimal performance on downstream tasks.

% KS: cite to scaling laws paper maybe on getting better performance with more layers/data?

This correspondence is probably (hopefully) trivial to most NLP researchers, but worth pointing out, lest we forget.

\section{Data generation}
\label{app:data}

We provide pseudo-code for $\mathcal{F}_i$ in Algorithm \autoref{algo:randomize}.
Following \citet{sinha2020b}, we do not explicitly control whether the permuted words maintain any of their original neighbors. Thus, a certain amount of extra n-grams are expected to co-occur, purely as a product of random shuffling. We quantify the amount of such shuffling on a sample of 1 million sentences drawn from the BookWiki random corpus, and present the BLEU-2, BLEU-3 and BLEU-4 scores in \autoref{tab:bleu_score}. We provide a sample snapshot of the generated data in \autoref{tab:sample_permutations}.

% Please add the following required packages to your document preamble:
% \usepackage{graphicx}
\begin{table}[]
\centering
\resizebox{\linewidth}{!}{%
\begin{tabular}{lrrr}
\toprule
 & BLEU-2 & BLEU-3 & BLEU-4 \\ \cline{2-4} 
\RI & 0.493 +/- 0.12 & 0.177 +/- 0.16 & 0.040 +/- 0.11 \\
\RII & 0.754 +/- 0.07 & 0.432 +/- 0.18 & 0.226 +/- 0.19 \\
\RIII & 0.824 +/- 0.06 & 0.650 +/- 0.09 & 0.405 +/- 0.20 \\
\RIV & 0.811 +/- 0.08 & 0.671 +/- 0.11 & 0.553 +/- 0.12 \\ \bottomrule
\end{tabular}%
}
\caption{BLEU-2,3,4 scores (mean and std dev) on a sample of 1M sentences drawn from the corpus used to train \RI, \RII, \RIII and \RIV\ compared to \OR.}
\label{tab:bleu_score}
\end{table}

\begin{algorithm}
\small
\caption{SentenceRandomizer}
\begin{algorithmic}[1]

\Procedure{$\mathcal{F}$}{$S,t,n$}       \Comment{Randomize a sentence $S$ with seed $t$ and n grams $n$}
    \State $W$ = tokenize the words in $S$
    \State Set the seed to $t$ 
    \If{$n > 1$}
        \While{True}
        \State $K$ = Sample all possible starting points from $[0, |W| - n]$
        \State Ignore the starting points in $K$ which overlap with conjoined tokens  \Comment{Conjoined tokens consists of joined unigrams}
        \If{$|K| \geq 1$}
        \State Sample one position $p \in K$
        \State $g$ = Extract the n-gram $W[p:p+n]$
        \State Delete $W[p+1:p+n]$
        \State $W[p]$ = Convert $g$ to a conjoined token 
        \Else
        \State Break from While loop
        \EndIf
        \EndWhile
    \EndIf
    \While{True}
        \State $\hat{W}$ = randomly shuffle tokens in $W$
        \State $r = \sum(\hat{W}[i] = W[i])$ \Comment{Count number of positions where the token remains in its original position}
        \If{$r = 0$}
            Break out of While loop
        \EndIf
    \EndWhile
    \State $\hat{S}$ = join the tokens in $\hat{W}$
    \State Return $\hat{S}$
\EndProcedure

\end{algorithmic}
\label{algo:randomize}
\end{algorithm}

% BLEU score and Perplexity

\begin{figure}[t]
    \centering
    \resizebox{0.48\textwidth}{!}{
        \includegraphics{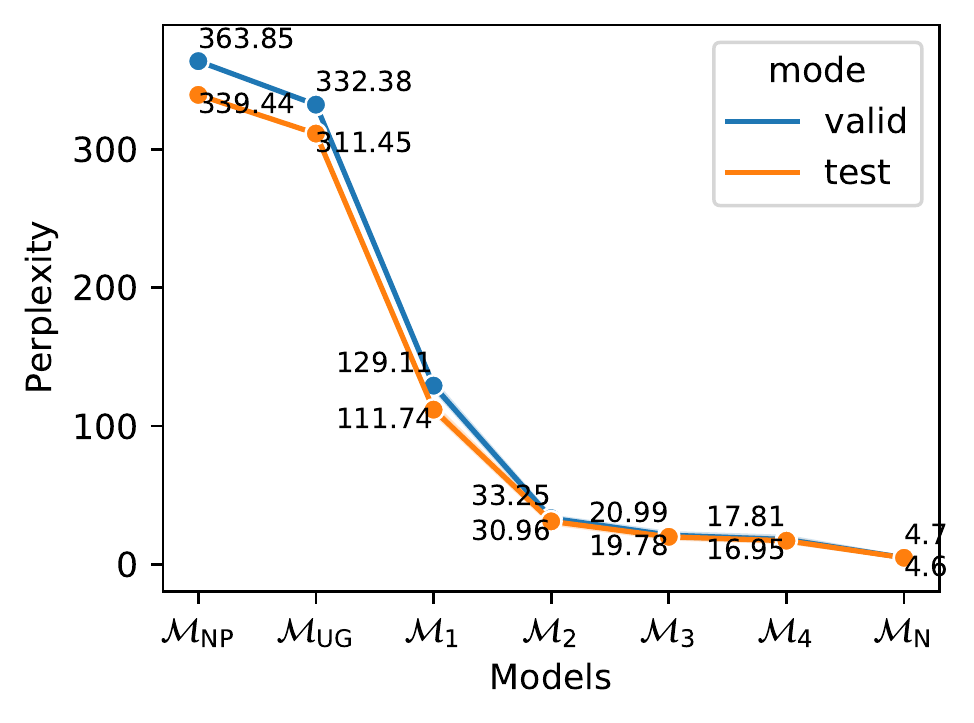}}
    \caption{Perplexity of various models on Wiki 103 valid and test sets.}
    \label{fig:perplexity}
\end{figure}

\section{Pre-training details}
\label{app:pretrain}

We use the Fairseq \cite{ott2019fairseq} toolkit to pre-train RoBERTa (base) models on the different variants of the BookWiki corpus. We follow the default parameters as reported in \citet{liu2019b}, with the following adjustments: max steps 100k, warmup steps: 20k.
We use the Wiki 103 validation and test set to validate and test the array of pre-trained models, as validation on this small dataset is quick, effective, and reproducible for comparison among publicly available datasets (\autoref{fig:perplexity}). We observe that perplexity monotonically increases from \OR, through \RIV--\RI, to \RC, and finally \NP.

% TODO: insert table of hyperparams

\section{Word-order pre-training ablations}
\label{app:ablations}

\begin{table*}[t]
  \centering
  \resizebox{\linewidth}{!}{%
\begin{tabular}{lllllllll}
\toprule
Model & QNLI & RTE & QQP & SST-2 & MRPC & PAWS & MNLI-m/mm & CoLA \\ \midrule
\OR & 92.45 +/- 0.2 & 73.62 +/- 3.1 & 91.25 +/- 0.1 & 93.75 +/- 0.4 & 89.09 +/- 0.9 & 94.49 +/- 0.2 & 86.08 +/- 0.2 / 85.4 +/- 0.2 & 52.45 +/- 21.2 \\ \midrule
\RIV & 91.65 +/- 0.1 & 70.94 +/- 1.2 & 91.39 +/- 0.1 & 92.46 +/- 0.3 & 86.90 +/- 0.3 & 94.26 +/- 0.2 & 83.79 +/- 0.2 / 83.94 +/- 0.3 & 35.25 +/- 32.2 \\ 
\RIII & 91.56 +/- 0.4 & 69.75 +/- 2.8 & 91.22 +/- 0.1 & 91.97 +/- 0.5 & 86.22 +/- 0.8 & 94.03 +/- 0.1 & 83.83 +/- 0.2 / 83.71 +/- 0.1 & 40.78 +/- 23.0 \\
\RII & 90.51 +/- 0.1 & 70.00 +/- 2.5 & 91.33 +/- 0.0 & 91.78 +/- 0.3 & 85.90 +/- 1.2 & 93.53 +/- 0.3 & 83.45 +/- 0.3 / 83.54 +/- 0.3 & 50.83 +/- 5.80 \\
\RI & 89.05 +/- 0.2 & 68.48 +/- 2.5 & 91.01 +/- 0.0 & 90.41 +/- 0.4 & 86.06 +/- 0.8 & 89.69 +/- 0.6 & 82.64 +/- 0.1 / 82.67 +/- 0.2 & 31.08 +/- 10.0 \\
\midrule
\RV & 84.97 +/- 0.3 & 56.09 +/- 0.6 & 90.15 +/- 0.1 & 86.11 +/- 0.7 & 79.41 +/- 0.6 & 77.3 +/- 12.63 & 77.58 +/- 0.3 / 77.89 +/- 0.4 & 12.54 +/- 5.57 \\
\NP & 77.59 +/- 0.3 & 54.78 +/- 2.2 & 87.78 +/- 0.4 & 83.21 +/- 0.6 & 72.78 +/- 1.6 & 57.22 +/- 1.2 & 63.35 +/- 0.4 / 63.63 +/- 0.2 & 2.37 +/- 3.20 \\
\RU & 77.69 +/- 0.4 & 53.84 +/- 0.6 & 85.92 +/- 0.1 & 84.00 +/- 0.6 & 71.35 +/- 0.8 & 58.43 +/- 0.3 & 72.10 +/- 0.4 / 72.58 +/- 0.4 & 8.89 +/- 1.40 \\
\RC & 66.94 +/- 9.2 & 53.70 +/- 1.0 & 85.57 +/- 0.1 & 83.17 +/- 1.5 & 70.57 +/- 0.7 & 58.59 +/- 0.3 & 71.93 +/- 0.2 / 71.33 +/- 0.5 & 0.92 +/- 2.10 \\
\RT & 62.17 +/- 0.4 & 52.97 +/- 0.2 & 81.53 +/- 0.2 & 82.0 +/- 0.7 & 70.32 +/- 1.5 & 56.62 +/- 0.0 & 65.70 +/- 0.2 / 65.75 +/- 0.3 & 8.06 +/- 1.60 \\

 \bottomrule
\end{tabular}}
\caption{GLUE and PAWS-Wiki dev set results on different ablations of the RoBERTa (base) models, trained on variants of the BookWiki corpus (with mean and std dev). The top row is the original model, the middle half contains the sentence randomization models, and the bottom half contains the ablations.}
\label{table:glue_ablations}
\end{table*}

\begin{figure}[t]
    \centering
    \resizebox{0.48\textwidth}{!}{
        \includegraphics{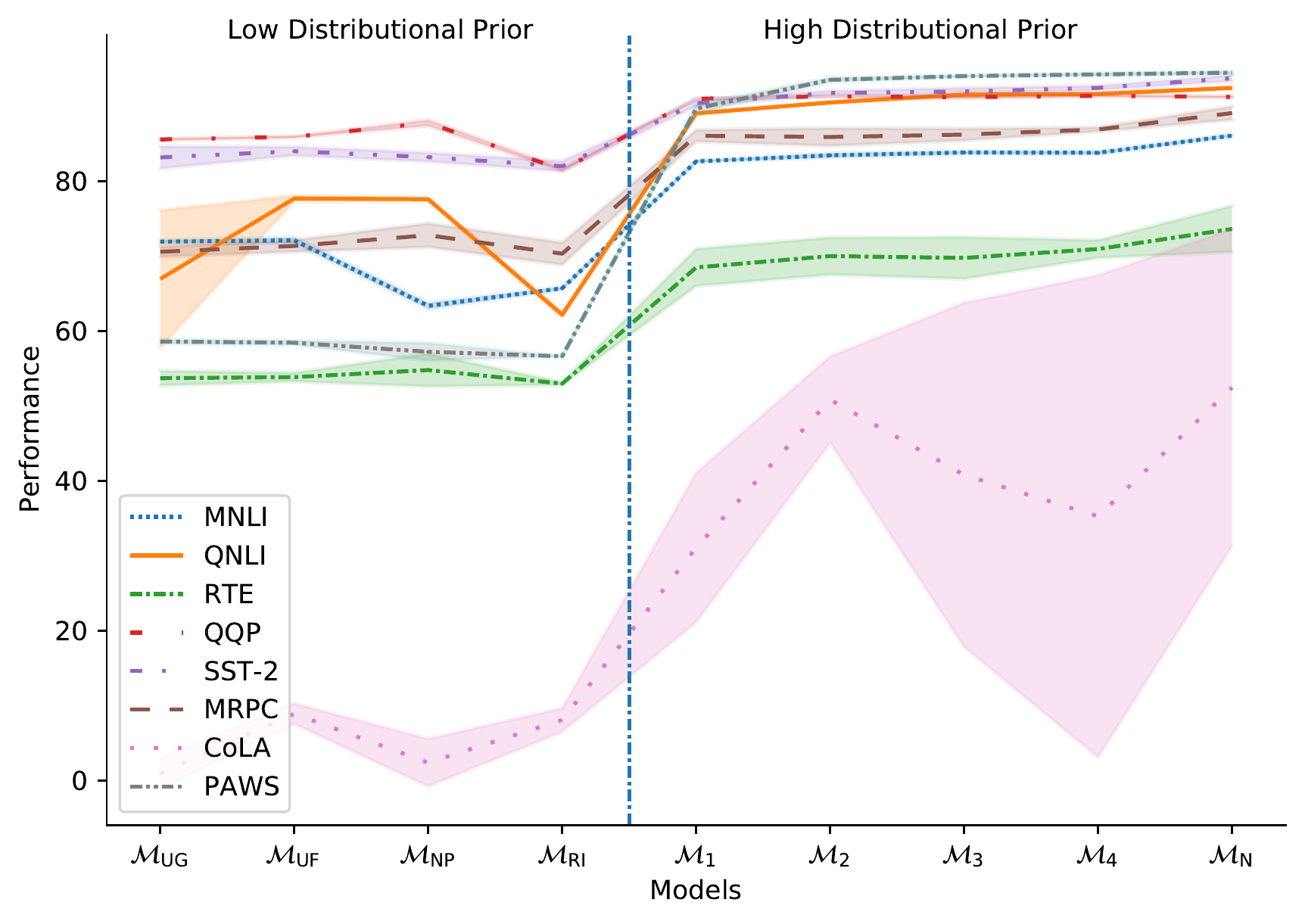}}
    \caption{GLUE results on various model ablations using BookWiki corpus.}
    \label{fig:all_glue_plot}
\end{figure}

We also train further model ablations with low to high distributional priors. Following the construction of the corpus bootstrap resample, we train a model where words are drawn uniformly from BookWiki corpus, thus destroying the natural frequency distribution (\RU). We further study an ablation for a high distributional prior, \RV, where we shuffle words (unigram) in a buffer created with joining multiple sentences such that maximum token length of the buffer is 512. This ablation---which is similar to the paragraph word shuffle condition in \newcite{gauthier-levy-2019-linking}---will allow us to study the effect of unigram shuffling in a window larger than the one for \RI. Buffer size is chosen to be 512 because BERT/RoBERTa is typically trained with that maximum sequence length.

We observe dev set results on the GLUE benchmark of these ablations, along with baselines \RC, \RT\ and \NP\ and random shuffles in \autoref{table:glue_ablations} and \autoref{fig:all_glue_plot}. We observe that \RV\ exhibits worse overall scores than \RI, however it is still significantly better than \NP\ or \RC\ baselines. We observe that destroying the natural frequency distribution of words (\RU) yields comparable or slightly better results compared to random corpus model \RC. This result shows that merely replicating the natural distribution of words without any context is not useful for the model to learn. These results indicate that at least some form of distributional prior is required for MLM-based models to learn a good downstream representation.

\begin{table}[t]
  \centering
  \resizebox{\linewidth}{!}{%
   \begin{tabular}{lrrrrrrrr}
\toprule
         Model &   RTE &  MRPC &  SST-2 &  CoLA &   QQP &  QNLI &  MNLI &  PAWS \\
\midrule
\RI & 68.48 & 85.97 &  90.41 & 31.07 & 91.01 & 89.05 & 82.64 & 89.69 \\
\RI$^*$ & 68.41 & 85.75 &  90.17 & 50.14 & 91.02 & 89.50 & 82.92 & 91.99 \\
\bottomrule
   \end{tabular}}
   \caption{Reconstruction experiments on shuffled word order sentences by fixing the same seed for every sentence (\RI) and having different seed for different shards of the corpus (\RI$^{*}$). We observe minimal difference in the downstream GLUE and PAWS scores.}
   \label{table:glue_recon}
\end{table}

One might argue that the superior results displayed by the unnatural models is due to the ability of RoBERTa to ``reconstruct'' the natural word order from shuffled sentences. The data generation algorithm, $\mathcal{F}_{i}$ requires a seed $t$ for every sentence. In our experiments, we had set the same seed for every sentence in the corpus to ensure reproducibility. However, it could be problematic if the sentences of the same length are permuted with the same seed, which could be easier for the model to ``reconstruct'' the natural word order to learn the necessary syntax. We tested this hypothesis by constructing a new corpus with different seeds for every sentence in every shard in the corpus (1/5th of BookWiki corpus is typically referred to as a \textit{shard} for computational purposes), to build the model \RI$^{*}$. We observe that there is minimal difference in the raw numbers among \RI\ and \RI$^{*}$ for most of the tasks (\autoref{table:glue_recon}) (with the exception of CoLA which performs similar to \RII\, possibly due to a difference in initialization). This result consequently proves that even with same seed, it is difficult for the model to just reconstruct the unnatural sentences during pre-training.

\section{Measuring Relative difference}
\label{app:rd}

In this section, we further measure the difference in downstream task performance reported in \autoref{subsec:glue_results} using as a metric the \textit{relative difference}. Let us denote the downstream task performance as $\mathcal{A}(\mathcal{T} | D)$, where $\mathcal{T}$ is the task and $D$ is the pre-trained model. We primarily aim to evaluate the relative performance gap, i.e. how much the performance differs between our natural and unnatural models. Thus, we define the \textit{Relative Difference} ($\Delta_{\{D\}}(\mathcal{T})$):

\begin{equation}
 \Delta_{\{D\}}(\mathcal{T}) =  \frac{\mathcal{A}(\mathcal{T} | OR) - \mathcal{A}(\mathcal{T} | D))}{\mathcal{A}(\mathcal{T} | OR) - \mathcal{A}(\mathcal{T} | \emptyset)},   
\end{equation}

where $\mathcal{A}(\mathcal{T} | \emptyset)$ is the random performance on the task $\mathcal{T}$ ($0.33$ for MNLI, $0$ for CoLA, and $0.5$ for rest)
$\Delta_{\{D\}}(\mathcal{T}) \rightarrow 0$ when the performance of a pre-trained model reaches that of the pre-trained model trained with natural word order.

We observe the relative difference on the tasks in \autoref{tab:glue_delta}. CoLA has the largest $\Delta_{\{D\}}(\mathcal{T})$ among all tasks, suggesting the expected high word order reliance. $\Delta_{\{D\}}(\mathcal{T})$ is lowest for QQP. %, which falls on the opposite end of the spectrum.

\begin{table}[t]
  \centering
  \resizebox{\linewidth}{!}{%
\begin{tabular}{lrrrrrrrr}
\toprule
Model &  QNLI &  RTE &  QQP &  SST-2 &  MRPC &  CoLA &  PAWS &  MNLI \\
\midrule
   \RI &    3.70 &   7.04 &   0.26 &     3.58 &    3.42 &   40.74 &    5.12 &  3.62 \\
   \RII &    2.11 &   4.95 &  -0.09 &     2.12 &    3.61 &    3.09 &    9.06 &  2.63 \\
   \RIII &    0.97 &   5.30 &   0.03 &     1.91 &    3.24 &   22.25 &    0.49 &  2.31 \\
   \RIV &    0.87 &   3.67 &  -0.15 &     1.39 &    2.47 &   32.79 &    0.25 &  2.19 \\\midrule
   \RC &   27.74 &  27.25 &   6.26 &    11.35 &   20.91 &   98.24 &   38.20 & 16.56 \\
   \NP &   16.16 &  25.77 &   3.83 &    11.30 &   18.42 &   95.48 &   39.66 & 26.10 \\
\bottomrule
\end{tabular}}
\caption{$ \Delta_{\{D_i\}}(\mathcal{T})$, scaled by a factor of 100 for GLUE and PAWS tasks.}
\label{tab:glue_delta}
\end{table}

\section{Fine-tuning with randomized data}
\label{app:rand_eval}

\begin{table*}[t]
\centering
\resizebox{\linewidth}{!}{%
\begin{tabular}{llllllllll}
\toprule
name & fine-tune-train & fine-tune-eval &            MNLI &            QNLI &             RTE &             CoLA &            MRPC &           SST-2 &             PAWS \\
\midrule
 \OR &        natural &       natural &  86.08 +/- 0.15 &  92.45 +/- 0.24 &  73.62 +/- 3.09 &  52.44 +/- 21.22 &  89.09 +/- 0.88 &  93.75 +/- 0.44 &   94.49 +/- 0.18 \\
  &        natural &           shuffled &  68.11 +/- 0.52 &  81.08 +/- 0.38 &  56.72 +/- 3.29 &    4.77 +/- 1.82 &  75.94 +/- 1.01 &  80.78 +/- 0.37 &   62.22 +/- 0.09 \\
  &            shuffled &       natural &  82.99 +/- 0.16 &  89.32 +/- 0.23 &   57.9 +/- 4.71 &      0.0 +/- 0.0 &  79.71 +/- 2.57 &   89.12 +/- 0.5 &  72.03 +/- 13.79 \\
  &            shuffled &           shuffled &   79.96 +/- 0.1 &  87.51 +/- 0.09 &   59.07 +/- 3.2 &     1.4 +/- 2.17 &  79.17 +/- 0.35 &   86.11 +/- 0.5 &   65.15 +/- 0.48 \\ \midrule
 \RI &        natural &       natural &  82.64 +/- 0.15 &  89.05 +/- 0.15 &  68.48 +/- 2.51 &   31.07 +/- 9.97 &  85.97 +/- 0.89 &  90.41 +/- 0.43 &   89.69 +/- 0.59 \\
  &        natural &           shuffled &  76.67 +/- 0.34 &  87.21 +/- 0.17 &   65.8 +/- 6.11 &    23.06 +/- 5.3 &  81.84 +/- 0.43 &  83.94 +/- 0.33 &   62.86 +/- 0.19 \\
  &            shuffled &       natural &   79.87 +/- 0.1 &  87.81 +/- 0.36 &  65.65 +/- 2.33 &  24.53 +/- 13.63 &  82.51 +/- 0.82 &  86.45 +/- 0.41 &   73.34 +/- 6.88 \\
  &            shuffled &           shuffled &   79.75 +/- 0.0 &  88.21 +/- 0.24 &  64.88 +/- 6.32 &  22.43 +/- 10.79 &  82.65 +/- 0.42 &   86.25 +/- 0.4 &    63.15 +/- 2.2 \\ \midrule
 \RC &        natural &       natural &  71.93 +/- 0.21 &  66.94 +/- 9.21 &   53.7 +/- 1.01 &    0.92 +/- 2.06 &  70.57 +/- 0.66 &   83.17 +/- 1.5 &   58.59 +/- 0.33 \\
  &        natural &           shuffled &  62.27 +/- 0.57 &  63.13 +/- 7.13 &  52.42 +/- 2.77 &    0.09 +/- 0.21 &  70.56 +/- 0.33 &  79.41 +/- 0.63 &   56.91 +/- 0.16 \\
  &            shuffled &       natural &   67.62 +/- 0.3 &  66.49 +/- 0.49 &  52.17 +/- 1.26 &      0.0 +/- 0.0 &  70.37 +/- 0.93 &  79.93 +/- 1.01 &   57.59 +/- 0.29 \\
  &            shuffled &           shuffled &  67.02 +/- 0.33 &  66.24 +/- 0.33 &  53.44 +/- 0.53 &    0.08 +/- 0.18 &  70.28 +/- 0.62 &   80.05 +/- 0.4 &   57.38 +/- 0.16 \\
\bottomrule
\end{tabular}
}
\caption{Fine-tuning evaluation by varying different sources of word order (with mean and std dev). We vary the word order contained in the pre-trained model (\OR,\RI,\RC); in fine-tuning training set (natural and shuffled); and in fine-tuning evaluation (natural and shuffled). Here, \textit{shuffled} corresponds to unigram shuffling of words in the input. In case of fine-tune evaluation containing shuffled input, we evaluate on a sample of 100 unigram permutations for each data point in the dev set of the corresponding task. }
\label{tab:full_eval}
\end{table*}

%TODO: rewrite this    

We perform additional experiments using the fine-tuned models from \autoref{subsec:glue_results}. Specifically, we construct unigram randomized train and test sets (denoted as \textit{shuffled}) of a subset of tasks to evaluate whether models fine-tuned on natural or unnatural task data (having natural or unnatural pre-training prior) are able to understand unnatural data during testing. \citeauthor{sinha2020b} showed for MNLI there exists at least one permutation for many examples which can be predicted correctly by the model. However, they also showed that every sentence can have many permutations which cannot be predicted correctly as well. We follow them in this evaluation, and construct 100 permutations for each example in the dev set for each task to capture the overall accuracy. 

Concretely, we use \OR, \RI\ and \RC\ as our pre-trained representations (trained with natural, unigram sentence shuffle and corpus shuffle data respectively) and evaluate the effect of training and evaluation on natural and unnatural data in \autoref{tab:full_eval}.
We observe that all models perform poorly on the \textit{shuffled} test set, compared to natural evaluation. However, interestingly, models have a slight advantage with a unigram randomized prior (\RI), with CoLA having the biggest performance gain.  PAWS task suffers the biggest drop in performance (from 94.49 to 62.22) but the lowest gain in \RI, confirming our conclusion from \autoref{subsec:glue_results} that most of the word order information necessary for PAWS is learned from the task itself.

Furthermore, training on shuffled data surprisingly leads to high performance on natural data for \OR\ in case of several tasks, the effect being weakest in case of CoLA and PAWS. This suggests that for tasks other than CoLA and PAWS, spurious correlations are leveraged by the models during fine-tuning (cf. \citealt{gururangan-etal-2018-annotation, poliak-etal-2018-hypothesis,tsuchiya-2018-performance}). We also observe evidence of \textit{domain matching}, where models improve their performance on evaluation on shuffled data when the training data source is changed from natural to shuffled (for \OR, MNLI shuffled evaluation improves from 68.11 to 79.96 just by changing the training corpus from natural to shuffled). We observe this behavior consistently for all tasks with all pre-trained representations.

\section{Dependency parsing using Second order Tree CRF Neural Dependency Parser}
\label{app:sota_dep}

\begin{table}[]
\centering
\resizebox{0.8\linewidth}{!}{%
\begin{tabular}{l|rl|ll}
\toprule
Model & \multicolumn{2}{c|}{UD EWT} & \multicolumn{2}{c}{PTB} \\ \hline
 & UAS & LAS & UAS & LAS \\ \cline{2-5} 
\OR &  90.92\% &  87.87\% & 95.42\% &  93.75\%  \\\midrule
\RI & 91.18\% &  88.19\% & 95.90\% &  94.35\% \\
\RII & 91.11\% &  88.12\% & 95.74\% &  94.16\% \\
\RIII & 91.05\% &  87.94\% & 95.73\% &  94.14\% \\
\RIV & 90.88\% &  87.78\% & 95.77\% &  94.16\% \\\midrule
\RC & 90.47\% &  87.42\% & 95.81\% &  94.28\% \\
\bottomrule
\end{tabular}%
}
\caption{Unlabeled Attachment Score (UAS) on Dependency parsing task on two datasets, UD EWT and PTB, using the Second order Tree CRF Neural Dependency Parser \cite{zhang-etal-2020-efficient}}
\label{tab:dependency_supar}
\end{table}

We also conduct extensive experiments with Second Order Tree CRF Neural Dependency parser from \citet{zhang-etal-2020-efficient}, using their provided codebase.\footnote{\href{https://github.com/yzhangcs/parser}{https://github.com/yzhangcs/parser}} We report the results on UD EWT and PTB corpus in \autoref{tab:dependency_supar}. Strangely enough, we find the gap to be even smaller across the different randomization models, even for some cases the performance on $R_1$ improves over $OR$. We suspect this result is due to two reasons: \textbf{(a)} Due to the presence of the complex Biaffine Dependency parser consisting of multiple LSTMs and individual MLP heads for each dependency arc (left and right), the majority of learning of the task is done by the parser itself; \textbf{(b)} \citet{zhang-etal-2020-efficient} downsample the BERT representation to 100 dimensions which is then combined with the learned LSTM representations, thereby minimizing the impact of the pre-trained representations. Our hypothesis is confirmed by the published results of \citet{zhang-etal-2020-efficient} on the Github repository, which shows a minimal gap between models with or without BERT.

\section{Perplexity analysis}

\begin{figure}[t]
    \centering
    \resizebox{0.5\textwidth}{!}{
        \includegraphics{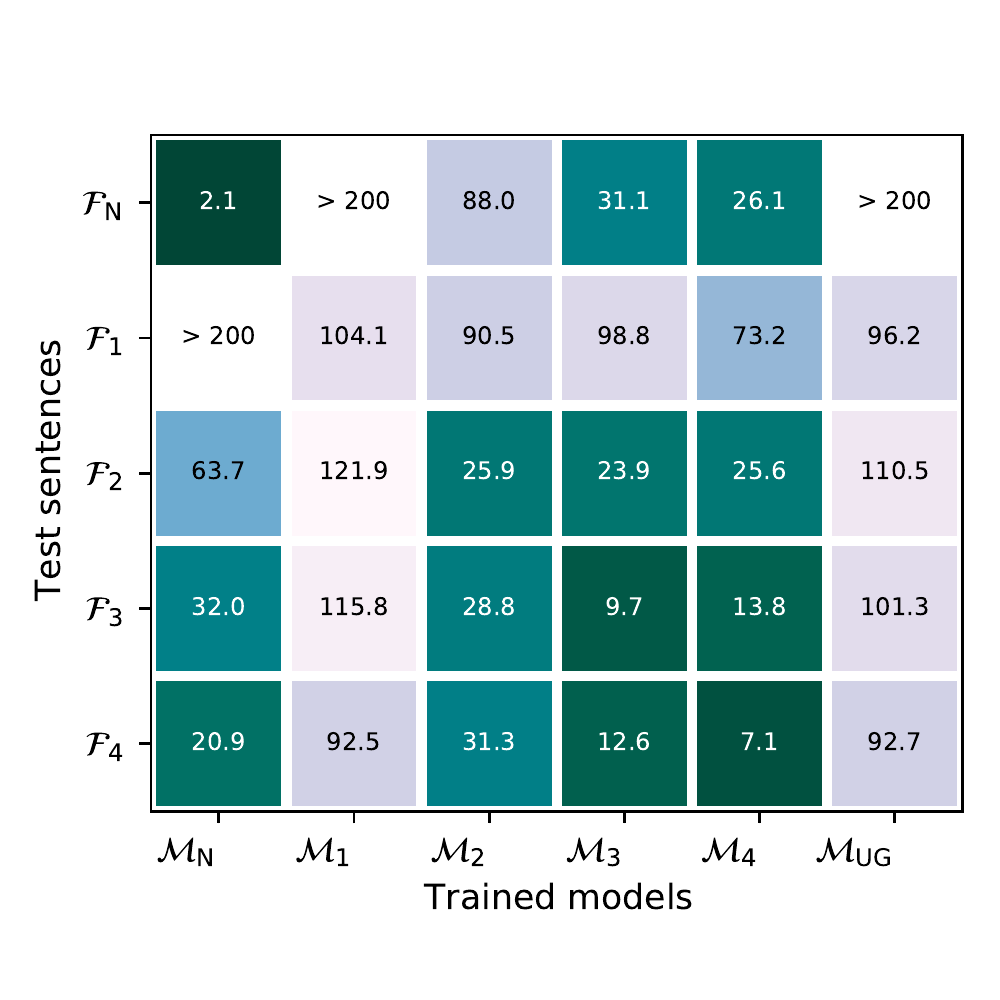}}
    \caption{BPPL scores per model per test scenario.}
    \label{fig:bpplm}
\end{figure}

We measure perplexity of various pre-trained randomization models on text that is randomized using the same function $\mathcal{F}$. Conventional language models compute the perplexity of a sentence $S$ by using past tokens ($S_{<t} = (w_1, w_2, \ldots, w_{t-1})$) and the application of chain rule ($\sum_{t=1}^{|S|} \log P_{\textit{LM}}(w_t | S_{t-1})$). However, this formulation is not defined for MLM, as a word is predicted using the entire sentence as a context.  Following \citet{salazar2020a}, we measure \textit{Pseudo-Perplexity}, i.e., given a sentence $S$, we compute the log-probability of the missing word in $S$ by iteratively masking out the specific word, and computing the average log-probability per word in $S$:

\begin{equation}
    \texttt{PLL}(S) = \frac{1}{|S|} \sum_{w \in S} \log P_{\texttt{MLM}} (w | S_{\setminus w}; \theta) 
\end{equation}

We bootstrap the \texttt{PLL} score of a test corpus $T$ by drawing 100 samples five times with replacement. We also similarly compute the bootstrap perplexity  following \citeauthor{salazar2020a}:

\begin{equation}
    \texttt{BPLL}_{T} = \exp( - \frac{1}{N} \sum_{S \in W} \texttt{PLL}(S)),
\end{equation}

where $W$ is the combined bootstrap sample containing $N$ sentences drawn with replacement from $T$. We compute this score on 6 pre-trained models, over four randomization schemes on the bootstrapped sample $W$ (i.e., we use the same n-gram randomization function $\mathcal{F}_i$). Thus, we obtain a 5x6 matrix of $\texttt{BPLL}$ scores, which we plot in \autoref{fig:bpplm}.

\begin{figure*}[ht]
    \centering
    \resizebox{\textwidth}{!}{
        \includegraphics{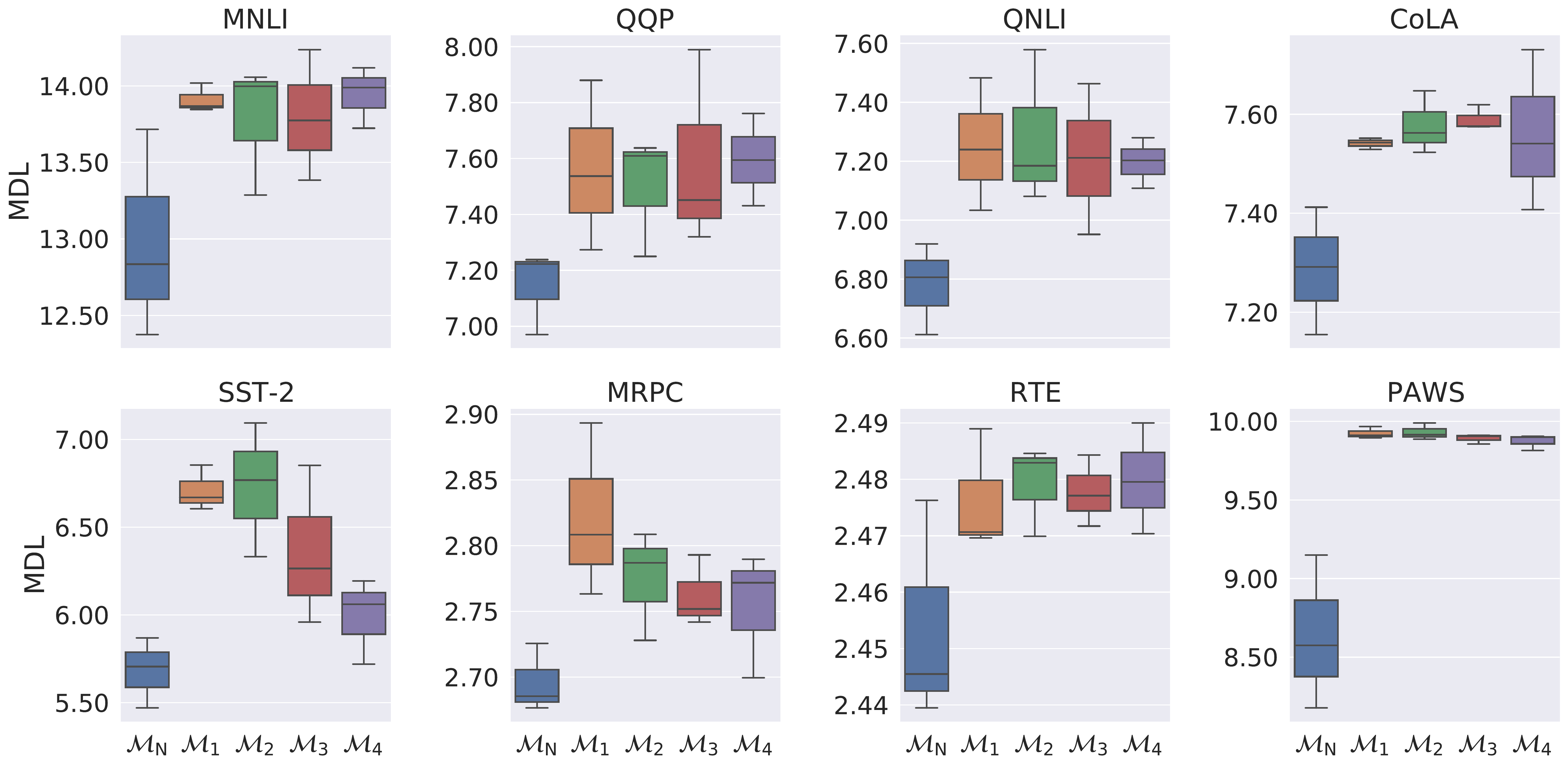}}
    \caption{Rissanen Data Analysis \cite{perez2021} on the GLUE benchmark and PAWS datasets. The lower minimum description length (MDL, measured in kilobits), the better the learning ability of the model.}
    \label{fig:rda}
\end{figure*}

We observe that the pre-trained model \OR\ has the lowest perplexity on the sentences with natural word order. Pre-trained models with random word order exhibit significantly higher perplexity than the normal word order sentences (top row). With the exception of \RI, the models pre-trained on randomized data (\RII, \RIII\ and \RIV) all display the lowest perplexity for their respective $n=2,3,4$ randomizations. %The unigram pre-trained model is a notable exception, which does not yield the lowest perplexity for $n=1$ randomization. 
These results indicate that the models  retain and detect the specific word order for which they were trained.  

%\subsection{Is word order in pre-trained representation helpful for better optimization during fine-tuning?}
\section{The usefulness of word order}
\label{app:rda_analysis}

The results in \autoref{subsec:glue_results} suggest that, with proper fine-tuning, an unnaturally trained model can reach a level of performance comparable to that of a naturally pre-trained model. However, we want to understand whether natural word order pre-training offers any advantage during the early stages of fine-tuning. Towards that end, we turn to compute the Minimum Description Length \cite[MDL; ][]{rissanen1984universal}. MDL is designed to characterize the complexity of data as the length of the shortest program required to generate it. Thus, the length of the minimum description (in bits) should provide a fair estimate of how much word order is useful for fine-tuning in a few-shot setting. Specifically, we leverage the Rissanen Data Analysis (RDA) framework from \citet{perez2021} to evaluate the MDL of pre-trained models on our set of downstream tasks. Under mild assumptions, if a pre-trained model $\theta_1$ is useful for solving a particular task $T$ over $\theta_2$, then the MDL in bits obtained by using $\theta_1$ should be shorter than $\theta_2$.
We follow the experimental setup of \citeauthor{perez2021} to compute the MDL on several tasks using $\theta$ = \{\OR,\RI,\RII,\RIII,\RIV\}, over three seeds and on three epochs of training. Concretely, RDA involves sampling 9 blocks of data from the dataset at random, where the size of each block is increased monotonically, training on 8 blocks while evaluating the model's loss (or \textit{codelength}) on the ninth. The minimum number of data samples in the smallest block is set at 64, while the largest number of data samples used in the last block is 10,000.

We observe that the value of MDL is consistently lowest for naturally pre-trained data (\autoref{fig:rda}). For purportedly word order reliant datasets such as RTE, CoLA and PAWS, the gap between the MDL scores among the natural and unnatural models is high. PAWS, specifically, has the largest advantage in the beginning of optimization, however with more fine-tuning, the model re-learns correct word order (\autoref{subsec:glue_results}). The present analyses, when taken in conjunction with our main results in \autoref{subsec:glue_results}, suggest that fine-tuning on large training datasets with complex classifiers in the pursuit of state-of-the-art results has mostly nullified the impact of word order in the pre-trained representations. Few shot \cite{bansal-etal-2020-learning} and few sample \cite{zhang2021a} learning and evaluation could potentially require more word order signal, thereby encouraging the model to leverage its own learned syntax better. 

% \subsection{Impact on downstream tasks: More data vs More words}

% Wiki Huge experiments

\section{At what point do models learn word order during pre-training?}
% KS: I kind of moved this to appendix as I think the reasoning/ explanation needs to be fleshed out more.

Results from \autoref{subsec:glue_results} beg the question: when, if at all, during pre-training does a model learn the natural word order? We aim to answer that question by comparing downstream task performance of RoBERTa base on intermediate checkpoints with that of the random word order pretrained models. The idea is to find the point during pre-training on natural corpus at which the model exceeds the task performance of the random pre-training model.

\begin{figure}[t]
    \centering
    \resizebox{0.5\textwidth}{!}{
        \includegraphics{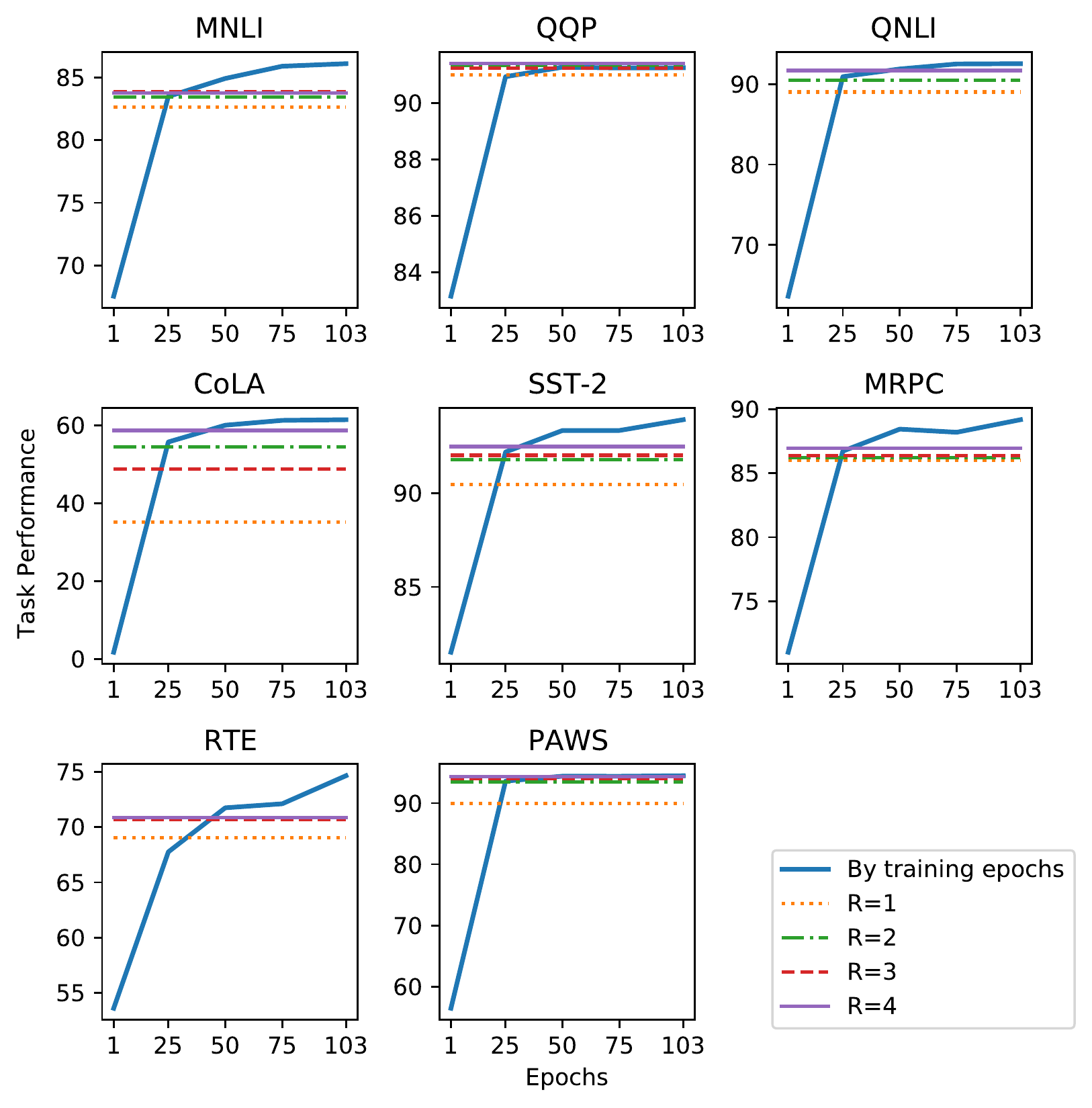}}
    \caption{Comparison among GLUE task performance from different steps in pre-training of RoBERTa on BookWiki Corpus.}
    \label{fig:glue_progression}
\end{figure}

Performance on all tasks (\autoref{fig:glue_progression}) increases rapidly during the first 20-25 epochs of pre-training. For some tasks, the word order information only helps after 30-50 pre-training epochs.

% \begin{itemize}
%     \item \autoref{fig:glue_progression} \textit{All tasks show rapid increase in performance during first 20-25 epochs of pre-training (~20-25k updates)}
%     \item \textit{For some tasks, the extra word order information helps in model accuracy after 50 epochs.}
% \end{itemize}

\section{More results from Syntactic Probes}
\label{app:pareto_probes}

\begin{table}[]
\centering
\resizebox{\linewidth}{!}{%
\begin{tabular}{l|rl|rl}
\toprule
Model & \multicolumn{2}{c|}{UD EWT} & \multicolumn{2}{c}{PTB} \\ \hline
 & MLP & Linear & MLP & Linear \\ \cline{2-5} 
\OR & 93.74 +/- 0.15 & 88.82 +/- 0.42 & 97.07 +/- 0.38 & 93.1 +/- 0.65 \\
\midrule
\RI & 88.60 +/- 3.43 & 80.76 +/- 3.38 & 95.33 +/- 0.37 & 87.83 +/- 1.86 \\
\RII & 93.39 +/- 0.45 & 87.58 +/- 1.06 & 96.96 +/- 0.15 & 91.80 +/- 0.50 \\
\RIII & 92.89 +/- 0.65 & 86.78 +/- 1.32 & 97.03 +/- 0.13 & 91.70 +/- 0.70 \\
\RIV & 92.83 +/- 0.61 & 87.23 +/- 0.77 & 96.96 +/- 0.12 & 92.08 +/- 0.39 \\\midrule
\RC & 89.10 +/- 0.21 & 79.75 +/- 0.5 & 94.12 +/- 0.01 & 84.15 +/- 0.51 \\ \bottomrule
\end{tabular}%%
}
\caption{Accuracy on the part-of-speech labelling task (POS) on two datasets, UD EWT and PTB, using the Pareto Probing framework \cite{pimentel-etal-2020-information}.}
\label{tab:pareto_pos_tag}
\end{table}

\begin{table}[t]
\centering
\resizebox{\linewidth}{!}{%
\begin{tabular}{l|rl|rl}
\toprule
Model & \multicolumn{2}{c|}{UD EWT} & \multicolumn{2}{c}{PTB} \\ %\hline
 & MLP & Linear & MLP & Linear \\ \cline{2-5} 
\OR & 89.63 +/- 0.60 & 84.35 +/- 0.78 & 93.96 +/- 0.63 & 88.35 +/- 1.00 \\
\midrule
\RI & 83.55 +/- 3.31 & 75.26 +/- 3.08 & 91.10 +/- 0.38 & 82.34 +/- 1.37 \\
\RII & 88.57 +/- 0.68 & 82.05 +/- 1.10 & 93.27 +/- 0.26 & 86.88 +/- 0.87 \\
\RIII & 88.69 +/- 1.09 & 82.37 +/- 1.26 & 93.46 +/- 0.29 & 87.12 +/- 0.72 \\
\RIV & 88.66 +/- 0.76 & 82.58 +/- 1.04 & 93.49 +/- 0.33 & 87.30 +/- 0.79 \\\midrule
\RC & 84.93 +/- 0.34 & 76.30 +/- 0.52 & 89.98 +/- 0.43 & 78.59 +/- 0.68 \\ \bottomrule
\end{tabular}%%
}
\caption{Accuracy on the dependency arc labelling task (DAL) on two datasets (with mean and std dev), UD EWT and PTB, using the Pareto Probing framework \cite{pimentel-etal-2020-pareto}.}
\label{tab:pareto_dep_label}
\end{table}

\begin{table}[]
\centering
\resizebox{0.7\linewidth}{!}{%
\begin{tabular}{l|l|l}
\toprule
Model & UD EWT & PTB \\ \hline
\OR & 0.528 +/- 0.01 & 0.682 +/- 0.01  \\\midrule
\RI & 0.489 +/- 0.03 & 0.648 +/- 0.01  \\
\RII & 0.529 +/- 0.00 & 0.681 +/- 0.01  \\
\RIII & 0.528 +/- 0.02 & 0.689 +/- 0.01  \\
\RIV & 0.525 +/- 0.00 & 0.683 +/- 0.01  \\\midrule
\RC & 0.510 +/- 0.01 & 0.640 +/- 0.05  \\
\bottomrule
\end{tabular}%
}
\caption{Pareto Hypervolume of dependency parsing task (DEP) on two datasets (with mean and std dev), UD EWT and PTB, using the Pareto Probing framework \cite{pimentel-etal-2020-information}.}
\label{tab:pareto_hypervolume}
\end{table}

%As suggested in \citet{pimentel-etal-2020-pareto}, w
We computed the Pareto Hypervolume on the dependency parsing task \citep{pimentel-etal-2020-pareto}. The Pareto Hypervolume is computed as the Area Under Curve (AUC) score over all hyperparameter runs, where the models are arranged based on their complexity. We observe minimal differences in the Pareto Hypervolumes (\autoref{tab:pareto_hypervolume}) among \OR\ and the randomization models for both datasets.

We also investigated two ``easy'' tasks, Part-of-Speech tagging (POS) and Dependency Arc Labeling (DAL) from the Pareto Probing framework.
For POS (\autoref{tab:pareto_pos_tag}) and DAL (\autoref{tab:pareto_dep_label}), since these tasks are simpler than DEP, the gap between \OR\ and unnaturally pre-trained models reduces even more drastically. The gap between \OR\ and \RI\ reduces to just 3.5 points on average for PTB in both POS and DAL.

\begin{figure}[h]
    \centering
    \resizebox{\linewidth}{!}{
        \includegraphics{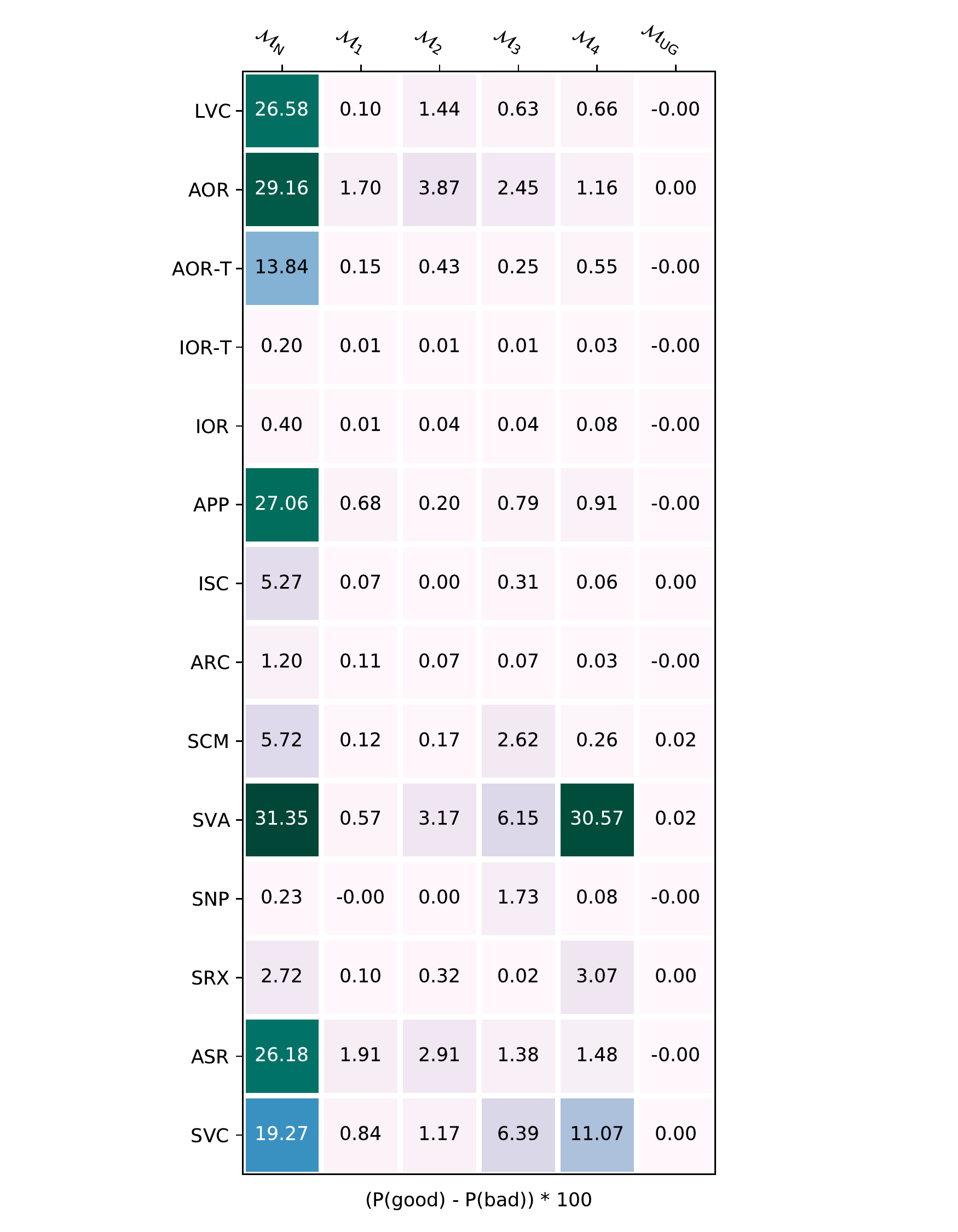}}
    \caption{The difference in word probabilities for stimuli in \citet{marvin-linzen-2018-targeted}: Simple Verb Agreement (SVA), In a sentential complement (SCM), Short VP Coordination (SVC), Long VP Coordination (LVC), Across a prepositional phrase (APP), Across a subject relative clause (ASR), Across an object relative clause (AOR), Across an object relative (no \textit{that}) (AOR-T),
    In an object relative clause (IOR),
    In an object relative clause (no \textit{that}) (IOR-T),
    Simple Reflexive (SRX), In a sentential complement (ISC),
    Across a relative clause (ARC), Simple NPI (SNP). }
    \label{fig:marv_diff}
\end{figure}

\section{Non parametric probes}
\label{app:non_par_probes}

% \begin{figure*}[ht]
%     \centering
%     \resizebox{\textwidth}{!}{
%         \includegraphics{fig/non_param_marv.pdf}}
%     \caption{\citet{marvin2018a}}
%     \label{fig:marvin_acc}
% \end{figure*}

% \begin{figure*}[ht]
%     \centering
%     \resizebox{\textwidth}{!}{
%         \includegraphics{fig/non_param_LGD.pdf}}
%     \caption{\citet{linzen-etal-2016-assessing}}
%     \label{fig:linzen_acc}
% \end{figure*}

% \begin{figure*}[ht]
%     \centering
%     \resizebox{\textwidth}{!}{
%         \includegraphics{fig/non_param_Gul.pdf}}
%     \caption{\citet{gulordava2018}}
%     \label{fig:gul_acc}
% \end{figure*}

\begin{figure}[]
    \centering
    \resizebox{\linewidth}{!}{
        \includegraphics{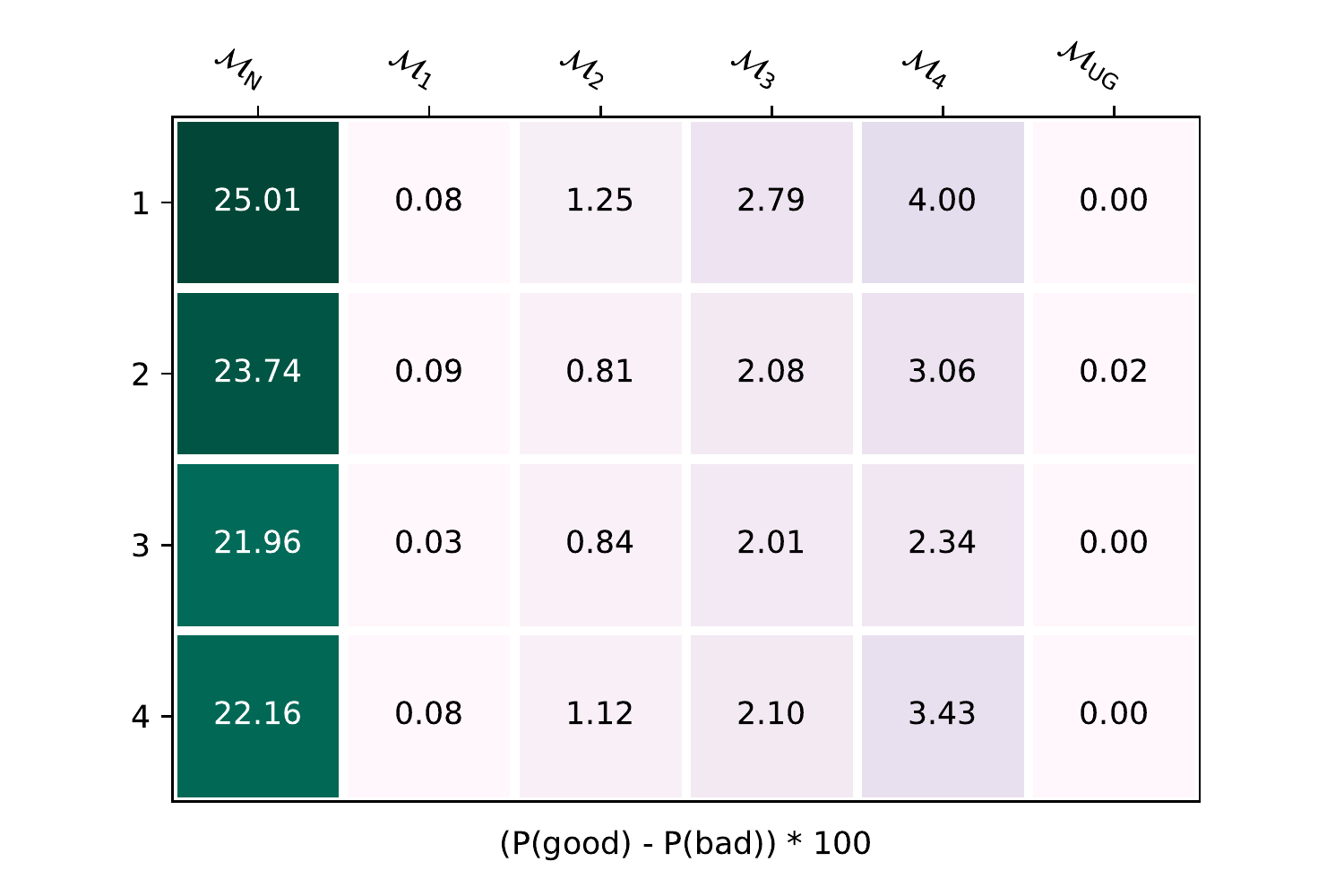}}
    \caption{\citet{linzen-etal-2016-assessing}}
    \label{fig:linzen_diff}
\end{figure}

\begin{figure}[h]
    \centering
    \resizebox{\linewidth}{!}{
        \includegraphics{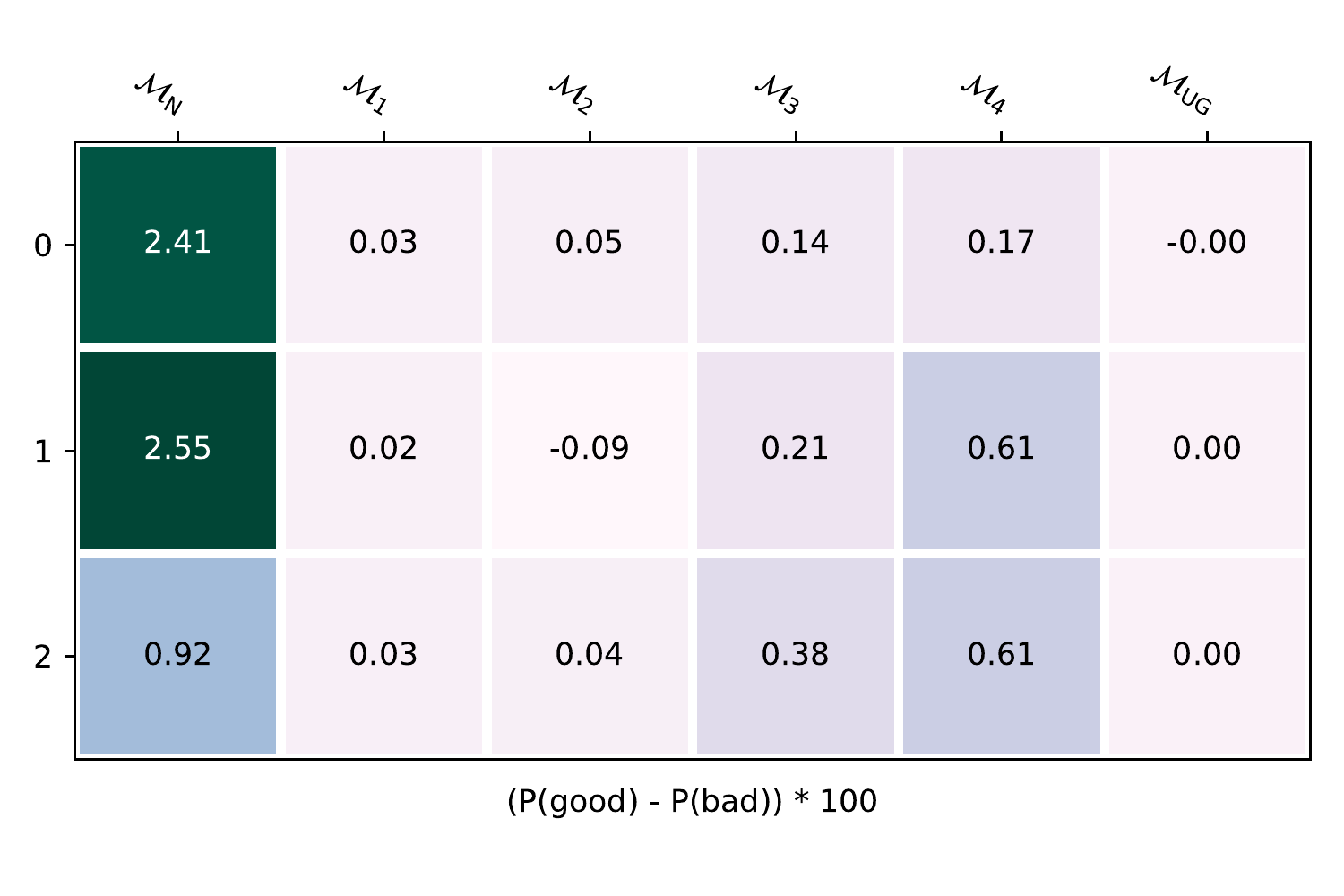}}
    \caption{\citet{gulordava2018}}
    \label{fig:gul_diff}
\end{figure}

\xhdr{Probability difference} In the original formulation \citep{goldberga, wolf2019}, the effectiveness of each stimulus is determined by the accuracy metric, computed as the number of times the probability of the correct focus word is greater than that of the incorrect word ($P(\textrm{good}) > P(\textrm{bad})$). We observed that this metric might not be reliable per se, since the probabilities may themselves be extremely low for all tokens, even when focus word probability decreases drastically from \OR\ to \RC.
Thus, we also report the mean difference of probabilities, $(\frac{1}{N}\sum_{i}^N P(\textrm{good}_i) - P(\textrm{bad}_i))$, scaled up by a factor of 100 for ease of observation, in \autoref{fig:gul_diff}, \autoref{fig:linzen_diff} and \autoref{fig:marv_diff}. 
We observe the highest difference between probabilities of the correct and incorrect focus words for the model pretrained on the natural word order (\OR). Moreover, with each step from \RI\ to \RIV, the difference between probablities of correct and incorrect focus words increases, albeit marginally, showing that pre-trained models with fewer n-gram words perturbed capture more word order information. \RC, the model with the distributional prior ablated, performs the worst, as expected. 

\begin{table*}
  \centering
  \resizebox{\linewidth}{!}{%
\begin{tabular}{lllllll}
\toprule
model & \OR & \RC & \RI & \RII & \RIII & \RIV \\
condition &               &                      &                   &                   &                   &                   \\
\midrule
1         &  93.45 (0.89) [25.04] &   58.87 (0.41) [0.0] &  59.96 (1.58) [0.08] &   63.63 (0.6) [1.25] &   64.7 (1.44) [2.79] &   70.47 (1.9) [4.01] \\
2         &    92.8 (1.22) [23.8] &  63.03 (1.35) [0.01] &   58.22 (1.5) [0.09] &  61.15 (2.07) [0.82] &  63.84 (2.41) [2.09] &   64.7 (1.92) [3.07] \\
3         &  87.71 (1.34) [22.03] &   64.06 (3.52) [0.0] &  56.69 (2.98) [0.03] &  56.83 (3.63) [0.85] &   61.1 (0.32) [2.02] &   63.0 (3.36) [2.35] \\
4         &  92.67 (0.52) [22.16] &   76.33 (1.38) [0.0] &  62.33 (7.61) [0.08] &  63.17 (9.09) [1.12] &   69.42 (1.77) [2.1] &  67.67 (7.02) [3.43] \\
\bottomrule
\end{tabular}}
\caption{\citet{linzen-etal-2016-assessing} stimuli results in raw accuracy. Values in parenthesis reflect the standard deviation over different seeds of pre-training. Values in square brackets indicate the mean probability difference among correct and incorrect words.}
\label{tab:linzen_full_balanced}
\end{table*}

\begin{table*}
  \centering
  \resizebox{\linewidth}{!}{%
\begin{tabular}{lllllll}
\toprule
model & \OR & \RC & \RI & \RII & \RIII & \RIV \\
condition &               &                      &                   &                   &                   &                   \\
\midrule
0         &   79.42 (5.5) [2.43] &  47.83 (3.76) [-0.0] &   53.67 (1.38) [0.03] &   58.75 (6.38) [0.05] &  63.58 (4.11) [0.14] &   63.75 (3.28) [0.17] \\
1         &  72.83 (4.07) [2.55] &     44.5 (0.5) [0.0] &    70.83 (5.8) [0.02] &  64.83 (1.76) [-0.09] &  71.67 (6.71) [0.21] &    71.5 (2.65) [0.61] \\
2         &   55.56 (0.0) [0.92] &  88.89 (11.11) [0.0] &  81.48 (12.83) [0.03] &   51.85 (6.42) [0.04] &  62.96 (6.42) [0.38] &  74.07 (16.97) [0.61] \\
\bottomrule
\end{tabular}}
\caption{\citet{gulordava2018} stimuli results in raw accuracy.Values in parenthesis reflect the standard deviation over different seeds of pre-training. Values in square brackets indicate the mean probability difference among correct and incorrect words.}
\label{tab:gul_full_balanced}
\end{table*}

\xhdr{Accuracy comparison}
We provide the accuracy as measured by \citet{goldberga, wolf2019} on the probing stimuli in \autoref{tab:linzen_full_balanced}, \autoref{tab:gul_full_balanced} and \autoref{tab:marvin_full_balanced}. We also highlight the difference in probability ($P(\textrm{good}) - P(\textrm{bad})$) in the table to provide a more accurate picture. All experiments were conducted on three pre-trained seeds for each model in our set of models. However, the low token probabilities in \RC\ tend to present unreliable scores. For example, in the case of \citet{gulordava2018} stimuli, unnatural models provide better scores compared to the natural model. We also observe for the \citet{linzen-etal-2016-assessing} stimuli that the results on model condition 4 (number of attractors) are surprisingly high for \RC\, whereas the individual token probabilities are lowest. We believe these inconsistencies stem from extremely low token probabilities themselves.  %\citet{gulordava2018} typically test on ``\textit{Colorless green ideas}" words, where the probability of a random word with the same POS tag is evaluated for the correct position.

\xhdr{Balancing datasets on inflection by upsampling}
The stimuli datasets of \citet{linzen-etal-2016-assessing} and \citet{gulordava2018} turned out to be heavily skewed towards words where singular was the correct inflection (as opposed to plural). This dataset imbalance caused the weak models (such as \RC) to have surprisingly high scores - the weak models were consistently providing higher probability for the singular inflection (\autoref{tab:linzen_full_unbalanced}). We upsample for both datasets, balancing the frequency of correct singular and plural inflections. We compute the upsampling number to the next multiple of 100 of the count of original singular inflections. For example, in condition 4 of \citet{linzen-etal-2016-assessing} dataset, we upsample both S and P to 300 rows each. This type of balancing via upsampling largely alleviated the inconsistencies we observed, and might prove to be useful when evaluating other models on these datasets in future.

\begin{table*}
  \centering
  \resizebox{\linewidth}{!}{%
\begin{tabular}{lllllll}
\toprule
Model &  \OR & \RC & \RI & \RII & \RIII & \RIV \\
condition &                &                      &                   &                   &                   &                   \\
\midrule
AOR       &  89.98 (1.96) [29.16] &    50.0 (0.01) [0.0] &    60.17 (1.61) [1.7] &    66.61 (7.1) [3.87] &  63.57 (2.39) [2.45] &   61.26 (4.91) [1.16] \\
AOR-T     &   77.4 (7.74) [13.84] &     50.0 (0.0) [0.0] &   78.88 (0.64) [0.15] &   52.17 (2.14) [0.43] &   48.85 (3.8) [0.25] &   57.06 (3.49) [0.55] \\
APP       &  89.94 (4.16) [27.06] &  50.01 (0.02) [-0.0] &    70.34 (1.9) [0.68] &     53.61 (3.3) [0.2] &  53.03 (1.75) [0.79] &    60.6 (4.41) [0.91] \\
ARC       &    85.06 (5.92) [1.2] &  50.05 (0.08) [-0.0] &   62.39 (1.91) [0.11] &   74.57 (5.99) [0.07] &  67.55 (3.84) [0.07] &   62.88 (3.45) [0.03] \\
ASR       &  87.19 (3.58) [26.18] &    50.0 (0.0) [-0.0] &  78.55 (10.01) [1.91] &    81.73 (5.1) [2.91] &   62.8 (0.35) [1.38] &   67.23 (6.82) [1.48] \\
IOR       &    89.83 (3.33) [0.4] &  50.55 (0.95) [-0.0] &   56.28 (2.66) [0.01] &   58.96 (4.28) [0.04] &   70.49 (2.2) [0.04] &   62.82 (8.51) [0.08] \\
IOR-T     &    74.05 (8.26) [0.2] &  50.61 (1.05) [-0.0] &   52.63 (2.07) [0.01] &   57.35 (4.88) [0.01] &  61.85 (4.75) [0.01] &   55.16 (6.59) [0.03] \\
ISC       &    85.87 (9.6) [5.27] &     50.0 (0.0) [0.0] &   67.85 (2.62) [0.07] &    82.66 (9.43) [0.0] &  77.69 (4.51) [0.31] &   68.65 (5.71) [0.06] \\
LVC       &   93.0 (0.75) [26.58] &  49.92 (0.14) [-0.0] &    70.42 (6.79) [0.1] &    87.5 (7.26) [1.44] &  85.42 (3.84) [0.63] &   81.08 (5.13) [0.66] \\
SCM       &    88.6 (3.49) [5.72] &    50.0 (0.0) [0.02] &   63.73 (7.94) [0.12] &   82.12 (0.92) [0.17] &  86.44 (3.67) [2.62] &   80.27 (2.46) [0.26] \\
SRX       &    91.0 (6.07) [2.72] &     50.0 (0.0) [0.0] &    88.0 (10.11) [0.1] &  92.25 (10.27) [0.32] &  94.25 (5.02) [0.02] &     91.0 (6.5) [3.07] \\
SVA       &  95.33 (7.23) [31.35] &    50.0 (0.0) [0.02] &    86.0 (5.29) [0.57] &  85.17 (12.87) [3.17] &  94.67 (5.25) [6.15] &  88.83 (9.57) [30.57] \\
SVC       &  97.54 (1.58) [19.27] &    50.0 (0.0) [-0.0] &   83.58 (4.58) [0.84] &   83.71 (8.78) [1.17] &   93.29 (7.4) [6.39] &  81.04 (3.66) [11.07] \\
\bottomrule
\end{tabular}}
\caption{\citet{marvin-linzen-2018-targeted} stimuli results in raw accuracy. Values in parenthesis reflect the standard deviation over different seeds of pre-training. Values in square brackets indicate the mean probability difference among correct and incorrect words. Abbreviations: Simple Verb Agreement (SVA), In a sentential complement (SCM), Short VP Coordination (SVC), Long VP Coordination (LVC), Across a prepositional phrase (APP), Across a subject relative clause (ASR), Across an object relative clause (AOR), Across an object relative (no \textit{that}) (AOR-T),
    In an object relative clause (IOR),
    In an object relative clause (no \textit{that}) (IOR-T),
    Simple Reflexive (SRX), In a sentential complement (ISC),
    Across a relative clause (ARC), Simple NPI (SNP).}
\label{tab:marvin_full_balanced}
\end{table*}

% \begin{table*}
%   \centering
%   \resizebox{\linewidth}{!}{%
% \begin{tabular}{lllllll}
% \toprule
% Model &               \OR &               \RI &               \RII &               \RIII &               \RIV &               \RC \\
% Condition &                  &                  &                  &                  &                  &                  \\
% \midrule
% 1         &  93.92 (2.5e+01) &   62.1 (7.8e-02) &  64.83 (1.3e+00) &  65.27 (2.8e+00) &  70.79 (4.0e+00) &  63.14 (1.6e-03) \\
% 2         &  93.02 (2.4e+01) &  62.86 (8.5e-02) &  62.75 (8.1e-01) &  65.08 (2.1e+00) &  65.44 (3.1e+00) &  71.98 (1.7e-02) \\
% 3         &  88.82 (2.2e+01) &  62.74 (3.4e-02) &  58.99 (8.4e-01) &  63.34 (2.0e+00) &  62.85 (2.3e+00) &  75.71 (3.2e-03) \\
% 4         &  90.53 (2.2e+01) &  63.16 (8.5e-02) &  63.94 (1.1e+00) &  66.41 (2.1e+00) &  66.28 (3.4e+00) &  80.54 (4.0e-03) \\
% \bottomrule
% \end{tabular}}
% \caption{\citet{linzen2016} stimuli results in raw accuracy. Values in parenthesis reflect the $P(\textrm{good}) - P(\textrm{bad})$ metric.}
% \label{tab:linzen_full}
% \end{table*}

\begin{table*}
  \centering
  \resizebox{\linewidth}{!}{%
\begin{tabular}{llllllll}
\toprule
Model &        \OR & \RC &   \RI &   \RII &  \RIII & \RIV & S/P \\
condition &                      &                      &                      &                      &                      &              &        \\
\midrule
1         &   94.04 (0.8) &          62.64 (0.5) &      62.18 (1.33) &      64.91 (0.14) &      65.35 (1.78) &      70.88 (1.88) & 14011 / 10112 \\
2         &  93.28 (0.94) &         71.24 (0.85) &      63.03 (1.69) &      62.92 (2.57) &      65.25 (3.13) &      65.61 (2.35) & 3120 / 1312 \\
3         &   89.1 (0.58) &         74.05 (1.85) &      62.94 (3.13) &      59.18 (3.32) &      63.54 (1.72) &       63.05 (2.0) & 733 / 215 \\
4         &   90.53 (0.9) &         80.03 (0.59) &      63.16 (4.83) &      63.94 (6.92) &      66.41 (3.17) &      66.28 (4.64) & 206 / 51 \\
\bottomrule
\end{tabular}}
\caption{\citet{linzen-etal-2016-assessing} stimuli results in raw accuracy on original, unbalanced data. Values in parenthesis reflect the standard deviation. S/P reflects the count of correct singular and plural focus words.}
\label{tab:linzen_full_unbalanced}
\end{table*}

% \input{tables/glue}

% Please add the following required packages to your document preamble:
% \usepackage{booktabs}
% \usepackage{graphicx}
\begin{table*}[]
\centering
\resizebox{\textwidth}{!}{%
\begin{tabular}{|l|p{0.3\linewidth}|p{0.3\linewidth}|p{0.3\linewidth}|p{0.3\linewidth}|p{0.3\linewidth}|}
\toprule
 & OR & R1 & R2 & R3 & R4 \\ \midrule
1 & They are commonly known as daturas, but also known as devil's trumpets, not to be confused with angel's trumpets, its closely related genus "Brugmansia". & be They angel's also but trumpets, genus related devil's as commonly closely known its daturas, trumpets, as "Brugmansia". confused with known are to not & as devil's They genus not to trumpets, closely related "Brugmansia". are commonly trumpets, its also known known as be confused daturas, but with angel's & "Brugmansia". related They are commonly trumpets, its closely as daturas, but known genus also known as trumpets, confused with angel's devil's not to be & its closely related genus They are commonly known trumpets, as trumpets, daturas, but also known as "Brugmansia". not to be confused with angel's devil's \\
2 & They are also sometimes called moonflowers, jimsonweed, devil's weed, hell's bells, thorn-apple, and many more. & are devil's bells, called weed, hell's thorn-apple, and many They also more. moonflowers, jimsonweed, sometimes & more. They are hell's bells, also sometimes and many called moonflowers, jimsonweed, devil's weed, thorn-apple, & jimsonweed, devil's weed, They are also thorn-apple, and many bells, more. hell's sometimes called moonflowers, & moonflowers, They are also sometimes bells, thorn-apple, and many more. called jimsonweed, devil's weed, hell's \\
3 & Its precise and natural distribution is uncertain, owing to its extensive cultivation and naturalization throughout the temperate and tropical regions of the globe. & throughout owing precise extensive temperate and naturalization and tropical of to natural is its Its distribution cultivation the globe. uncertain, regions the and & and natural distribution is tropical to its and naturalization throughout the the temperate and globe. Its precise uncertain, owing extensive cultivation regions of & uncertain, owing to Its precise and its extensive cultivation of globe. natural distribution is the the and tropical regions and naturalization throughout temperate & globe. Its precise and natural cultivation distribution the is uncertain, owing to its extensive and naturalization throughout the temperate and tropical regions of \\
4 & Its distribution within the Americas and North Africa, however, is most likely restricted to the United States, Mexico and Southern Canada in North America, and Tunisia in Africa where the highest species diversity occurs. & distribution Mexico occurs. likely diversity North however, species most the Tunisia where in and and North Canada Southern America, highest Africa United the and in Americas Its within States, is to the restricted Africa, & and Tunisia the Americas distribution within Mexico and is most United States, Africa, however, Africa where in North Its and North in Southern Canada America, the to the likely restricted occurs. highest species diversity & likely Its highest species diversity United States, Mexico restricted to the Africa where the occurs. distribution within the and Tunisia in however, is most Americas and Southern Canada and North Africa, in North America, & Tunisia occurs. Its distribution within the Africa where the highest in restricted to the United Canada in North America, most North Africa, however, is and Americas likely diversity States, Mexico and Southern species and \\
5 & All species of "Datura" are poisonous, especially their seeds and flowers. & seeds and species of poisonous, "Datura" their are All flowers. especially & "Datura" are especially their flowers. seeds and of All species poisonous, & especially their seeds flowers. "Datura" are poisonous, All species of and & flowers. poisonous, species of "Datura" are All especially their seeds and \\
6 & Some South American plants formerly thought of as "Datura" are now treated as belonging to the distinct genus "Brugmansia" ("Brugmansia" differs from "Datura" in that it is woody, making shrubs or small trees, and it has pendulous flowers, rather than erect ones). & and "Datura" treated from than flowers, it small belonging woody, thought as ones). South differs Some "Brugmansia" American as are in the rather pendulous distinct making now erect "Datura" to ("Brugmansia" of formerly trees, or is it that plants genus has shrubs & "Brugmansia" ("Brugmansia" than erect pendulous genus and ones). is woody, small trees, of as the distinct flowers, rather Some South differs from American plants treated as formerly thought belonging to "Datura" in making that it "Datura" are it has now shrubs or & woody, small trees, and has pendulous flowers, as belonging to Some making shrubs or as rather than erect "Datura" are now "Brugmansia" ("Brugmansia" differs the distinct genus from "Datura" in formerly thought of it treated that it is ones). South American plants & belonging to the distinct has making Some ("Brugmansia" differs from "Datura" in are now treated as genus pendulous shrubs flowers, rather than erect or ones). "Brugmansia" that it is woody, South American plants formerly thought of as "Datura" small trees, and it \\
7 & Other related taxa include & taxa Other include related & include Other related taxa & include Other related taxa & Other related taxa include \\
8 & "Hyoscyamus niger", "Atropa belladonna", "Mandragora officinarum", Physalis, and many more. & and many niger", officinarum", belladonna", "Mandragora "Atropa "Hyoscyamus more. Physalis, & belladonna", "Mandragora "Hyoscyamus niger", many Physalis, and more. officinarum", "Atropa & more. Physalis, and many belladonna", "Mandragora officinarum", "Hyoscyamus niger", "Atropa & niger", more. belladonna", "Mandragora officinarum", Physalis, "Atropa many and "Hyoscyamus \\
9 & The name "Datura" is taken from Sanskrit ' 'thorn-apple', ultimately from Sanskrit ' 'white thorn-apple' (referring to "Datura metel" of Asia). & of Asia). taken from name The "Datura" ' is to 'thorn-apple', Sanskrit ' Sanskrit metel" 'white (referring from "Datura thorn-apple' ultimately & "Datura" is taken from to ' 'thorn-apple', Sanskrit ' 'white of thorn-apple' (referring Asia). The name Sanskrit ultimately from "Datura metel" & Sanskrit ' The name "Datura" 'thorn-apple', ultimately from metel" Asia). is taken from of 'white (referring to "Datura Sanskrit ' thorn-apple' & Asia). The name "Datura" is from taken of from Sanskrit ' 'thorn-apple', ultimately Sanskrit ' 'white thorn-apple' (referring to "Datura metel" \\
10 & In the Ayurvedic text Sushruta different species of Datura are also referred to as ' and '. & the of also Sushruta Datura are referred to as In Ayurvedic and different species ' text '. & species of referred to are also Datura Sushruta different and as ' Ayurvedic text In the '. & as ' and In the Ayurvedic also referred to species of Datura are text Sushruta different '. & different In the Ayurvedic text also referred to as and Sushruta ' species of Datura are '. \\ \bottomrule
\end{tabular}%
}
\caption{First 10 lines from the BookWiki corpus, and their respective n-gram permutations.}
\label{tab:sample_permutations}
\end{table*}

\begin{table*}[t]
  \centering
  \resizebox{0.7\linewidth}{!}{%
\begin{tabular}{lrrrrrrrr}
\toprule
        Model &   RTE &  MRPC &  SST-2 &  CoLA &   QQP &  QNLI &  MNLI &  PAWS \\
\midrule
        \OR & 2e-05 & 2e-05 &  1e-05 & 2e-05 & 1e-05 & 1e-05 & 1e-05 & 2e-05 \\
         \RI & 2e-05 & 1e-05 &  1e-05 & 1e-05 & 3e-05 & 1e-05 & 2e-05 & 2e-05 \\
         \RII & 2e-05 & 2e-05 &  1e-05 & 1e-05 & 2e-05 & 1e-05 & 1e-05 & 3e-05 \\
         \RIII & 3e-05 & 1e-05 &  2e-05 & 2e-05 & 3e-05 & 1e-05 & 1e-05 & 2e-05 \\
         \RIV & 3e-05 & 1e-05 &  2e-05 & 2e-05 & 2e-05 & 1e-05 & 1e-05 & 2e-05 \\
         \RV & 1e-05 & 3e-05 &  2e-05 & 2e-05 & 3e-05 & 2e-05 & 3e-05 & 2e-05 \\
 \RC & 2e-05 & 1e-05 &  3e-05 & 1e-05 & 3e-05 & 3e-05 & 3e-05 & 2e-05 \\
\RU & 2e-05 & 1e-05 &  3e-05 & 2e-05 & 3e-05 & 3e-05 & 3e-05 & 1e-05 \\
   \RT & 1e-05 & 1e-05 &  3e-05 & 1e-05 & 1e-05 & 1e-05 & 2e-05 & 1e-05 \\
      \NP & 1e-05 & 3e-05 &  2e-05 & 1e-05 & 1e-05 & 1e-05 & 1e-05 & 1e-05 \\
\bottomrule
\end{tabular}}
\caption{Fine-tuning hyperparam Learning rate of each model for each task in GLUE and PAWS}
\label{table:glue_hyp_lr}
\end{table*}

\begin{table*}[t]
  \centering
  \resizebox{0.7\linewidth}{!}{%
\begin{tabular}{lrrrrrrrr}
\toprule
        Model &  RTE &  MRPC &  SST-2 &  CoLA &  QQP &  QNLI &  MNLI &  PAWS \\
\midrule
        \OR &   16 &    16 &     32 &    16 &   16 &    32 &    32 &    16 \\
         \RI &   32 &    32 &     16 &    32 &   32 &    16 &    32 &    16 \\
         \RII &   32 &    16 &     32 &    16 &   32 &    32 &    16 &    32 \\
         \RIII &   32 &    32 &     16 &    32 &   32 &    16 &    32 &    32 \\
         \RIV &   32 &    16 &     32 &    16 &   32 &    32 &    32 &    32 \\
        \RV &   32 &    16 &     16 &    32 &   32 &    16 &    16 &    16 \\
 \RC &   16 &    16 &     16 &    16 &   32 &    16 &    16 &    32 \\
\RU &   16 &    32 &     16 &    16 &   32 &    16 &    16 &    16 \\
   \RT &   16 &    16 &     32 &    16 &   16 &    16 &    32 &    16 \\
      \NP &   16 &    32 &     16 &    16 &   32 &    16 &    16 &    16 \\
\bottomrule
\end{tabular}
}
\caption{Finetuning hyperparam batch size of each model for each task in GLUE and PAWS}
\label{table:glue_hyp_bs}
\end{table*}

\end{document}